\documentclass[10pt,conference,compsoc]{IEEEtran}

\ifCLASSOPTIONcompsoc
  \usepackage[nocompress]{cite}
\else
  \usepackage{cite}
\fi

\ifCLASSINFOpdf
\else
\fi

\usepackage{cite}
\usepackage{amsmath,amssymb,amsfonts}
\usepackage{subcaption}
\usepackage{graphicx}
\usepackage{comment}
\usepackage[normalem]{ulem}
\usepackage{textcomp}
\usepackage{xcolor}
\usepackage{algorithm}%
\usepackage{soul}
\usepackage{algpseudocode}
\def\BibTeX{{\rm B\kern-.05em{\sc i\kern-.025em b}\kern-.08em
    T\kern-.1667em\lower.7ex\hbox{E}\kern-.125emX}}

\usepackage[normalem]{ulem}

\newcommand{\sai}[1]{{\textcolor{black}{ #1}}}
\newcommand{\zz}[1]{{\textcolor{black}{ #1}}}
\newcommand{\pstwo}[1]{{\textcolor{black}{ #1}}}
\newcommand{\zztbd}[1]{{\textcolor{black}{ #1}}}

\newcommand{\vv}[1]{{\textcolor{black}{ #1}}}

\algblock{ParFor}{EndParFor}
\algnewcommand\algorithmicparfor{\textbf{in parallel for}}
\algnewcommand\algorithmicpardo{\textbf{do}}
\algnewcommand\algorithmicendparfor{\textbf{end\ for}}
\algrenewtext{ParFor}[1]{\algorithmicparfor\ #1\ \algorithmicpardo}
\algrenewtext{EndParFor}{\algorithmicendparfor}
\algnewcommand\And{\textbf{and}}

\begin{document}
\title{
\huge{Privacy and Efficiency of Communications in \\ Federated Split Learning
}
}

\author{
\begin{tabular}{ccccc}

Zongshun~Zhang$^{\ddagger}$ & Andrea~Pinto$^{\star}$ & Valeria~Turina$^{\star}$ & Flavio Esposito$^{\star}$ &  Ibrahim Matta$^{\ddagger}$ \\

  \multicolumn{5}{c}{\normalsize{}} \\
  \multicolumn{2}{c}{$^{\ddagger}$\normalsize{Computer Science Department}} &  & \multicolumn{2}{c}{$^{\star}$\normalsize{Computer Science Department}} \\
  \multicolumn{2}{c}{\normalsize{Boston University}} &  & \multicolumn{2}{c}{\normalsize{Saint Louis University}} \\

  \multicolumn{5}{c}{{\tt Technical Report}} \\

\end{tabular}
}

\IEEEtitleabstractindextext{%
\begin{abstract}
Everyday, large amounts of sensitive data \sai{is} distributed across mobile phones, wearable devices, and other sensors. Traditionally, these enormous datasets have been processed on a single system, with complex models being trained to make valuable predictions.
Distributed machine learning techniques such as Federated and Split Learning have recently been developed to protect user \sai{data and} privacy better while ensuring high performance. Both of these distributed learning architectures have advantages and disadvantages.
In this paper, we examine these tradeoffs and suggest a new hybrid Federated Split Learning architecture that combines the efficiency and privacy benefits of both.
Our evaluation demonstrates how our hybrid Federated Split Learning approach can lower the amount of processing power required by each client running a distributed learning system, reduce training and inference time while keeping a similar accuracy. We also discuss the resiliency of our approach to deep learning privacy inference attacks and compare our solution to other recently proposed benchmarks.
\end{abstract}

}

\maketitle

\IEEEdisplaynontitleabstractindextext

%
\IEEEpeerreviewmaketitle

\section{Introduction}

\IEEEPARstart{C}{ENTRALIZED} machine learning (ML) training is becoming unsustainable~\cite{survey-on-FederatedLearning2022}. Aside from the advantages of re-training often to optimize revenues~\cite{ray}, several learning applications need to run their processes at the edge of the network, not in the core of a datacenter, for multiple reasons, including end-to-end latency minimization by running machine learning algorithms locally on an end-device, and privacy concerns of trusting third-party clouds~\cite{privacy-attack-toNN}.
Several Machine Learning (ML) models trade user experience improvements on mobile devices for sensible data exploitation; see e.g., text recommendation in keyboards~\cite{48270, 47586} or vocal assistants~\cite{KeywordSpottingFederated}.
In these and other applications, a decentralized learning approach may be preferable to a centralized system since sensitive data may remain locally within a client and not transferred over a computer network. 

Despite its benefits in several use cases, running machine learning training and inference jobs within local devices has several limitations: computing capacity is often limited, battery drains faster with intensive processing  and the mobile or other end-devices have limited memory and storage capabilities. 
For example, our experiments show that to fine-tune a VGG-16~\cite{VGG16} Neural Network, pre-trained on
ImageNet~\cite{ImageNet} with Cifar-10~\cite{cifar10}, tens of minutes are needed to reach $90\%$ accuracy on an NVDIA V100 GPU.

Different distributed neural network architectures have been proposed to preserve privacy and guarantee timely convergence -- for example Federated Learning~\cite{survey-on-FederatedLearning2022}, Split Learning~\cite{splitNN}, or hybrid approaches~\cite{FedSL,splitfed,parallelSplit}. 
Federated Learning (FL)~\cite{McMahan2017CommunicationEfficientLO} averages the weights of the learned Neural Network model on each edge device to create a single model, which will update the local ones.
Previous research has shown that this strategy can achieve higher accuracy than considering only a local model~\cite{McMahan2017CommunicationEfficientLO ,FedAvgProof-1, FedAvgProof-2-Free-rider, FedAvgProof-3-FedProx} and at the same time can preserve the privacy of the data.
\zz{Split Learning (SL) architecture splits the entire NN into partitions of layers. Each partition is executed on a different entity (i.e., edge and cloud), and different edge NN partitions can be paired with the cloud partition.}
Thus, this approach takes advantage of distributed datasets while keeping user data private.  

On one hand, FL is easy to scale to many devices given that there are enough resources to meet training Service Level Agreements (SLAs).
Thus, FL is impractical in edge  training/inference settings, where resources are limited.
On the other hand, Split Learning can train with limited resources, but it doesn't scale to many devices well since it is not parallel.
Especially when we pair different edge devices with not independent and identically distributed (Non-IID) data with the central cloud, the training may not converge at all~\cite{non-iid-distributed-learning}.
%
Another drawback of SL is that the intermediate data can be costly to transmit and store in client or server nodes~\cite{BBNet,bottlenet,early-exit,sl-distill}. 
Furthermore, since the intermediate data in the forward propagation is derived from the source data, there is a privacy concern.
We further discuss these two models in Section~\ref{background-and-related}.


To cope with the inefficiencies of the existing distributed learning models, 
\zz{we propose} a novel distributed learning architecture, Federated Split Learning  (FSL)~\cite{FSL}, which combines the benefits of FL and SL while mitigating their drawbacks.  
We discuss the generality of our FSL and a novel methodology to optimize both delay and privacy guarantees.

\begin{figure*}[!ht]
\captionsetup[subfigure]{justification=centering}
    \centering
    \begin{subfigure}[t]{0.24\textwidth}
    \includegraphics[width=\textwidth]{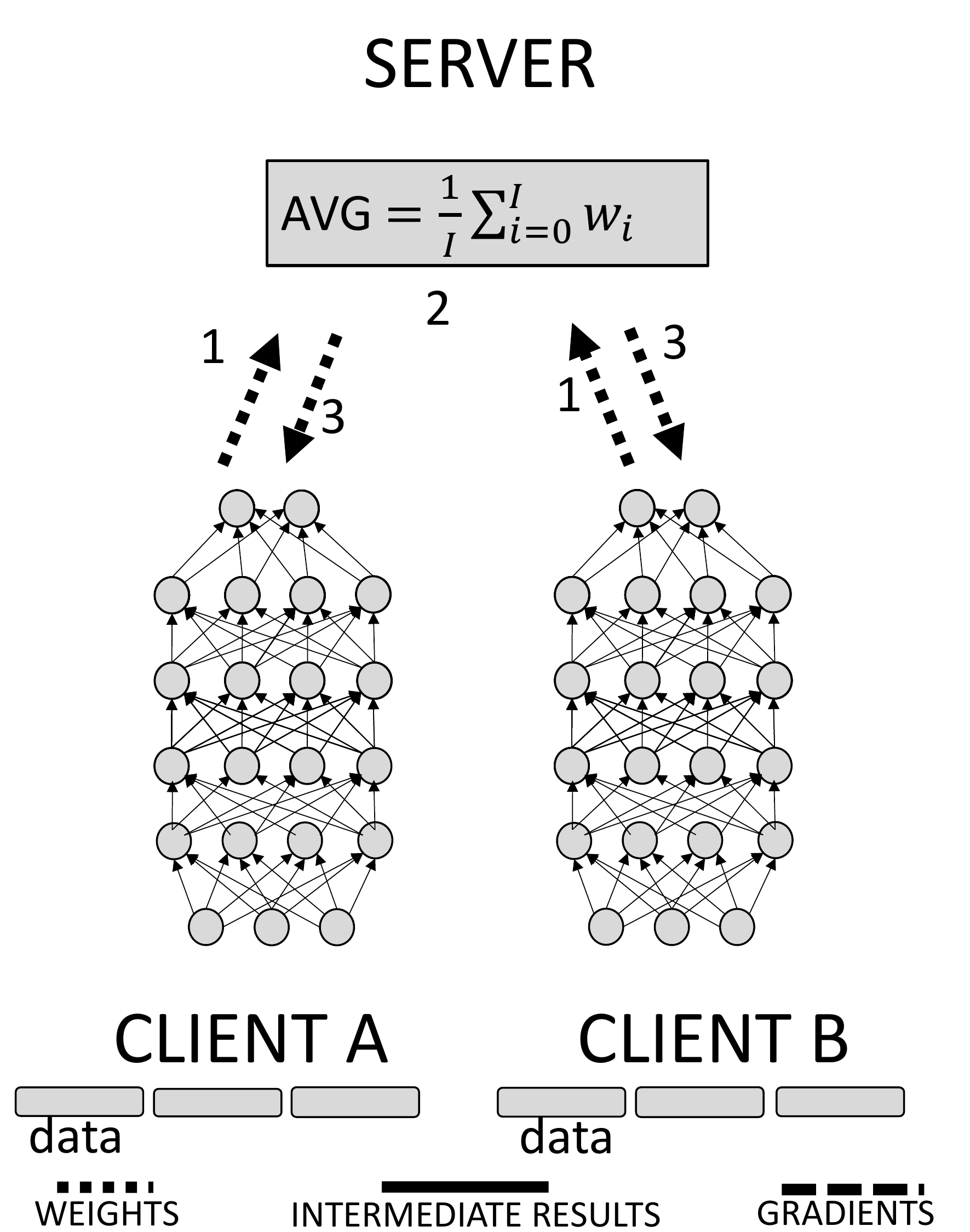}
    \caption{Federated Learning}
    \label{FL_arch_fig}
    \end{subfigure}
    \begin{subfigure}[t]{0.24\textwidth}
    \includegraphics[width=\textwidth]{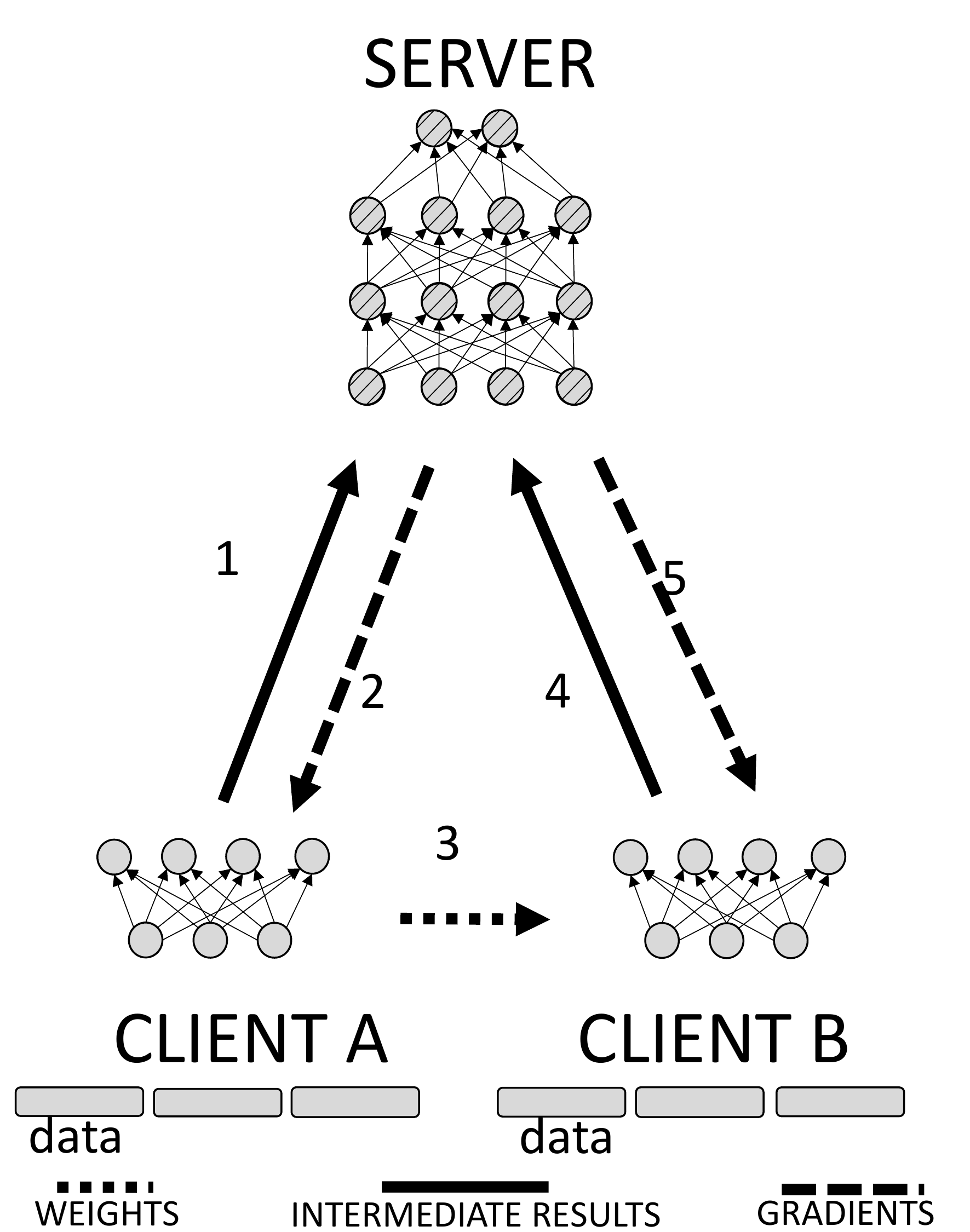}
    \caption{Split Learning}
    \label{SL_arch_fig}
    \end{subfigure}
    \begin{subfigure}[t]{0.24\textwidth}
    \includegraphics[width=\textwidth]{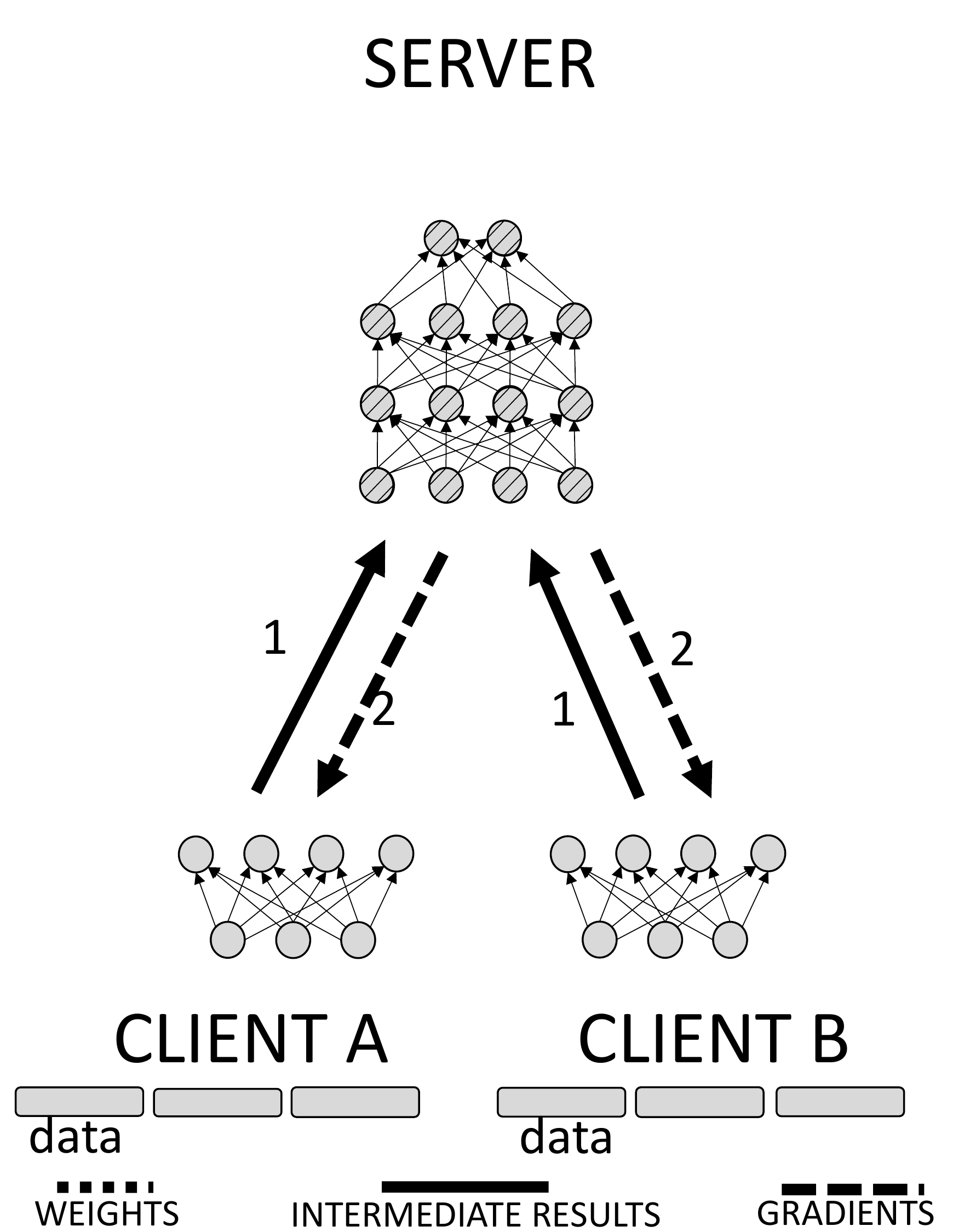}
    \caption{Parallel Split Learning}
    \label{PSL_arch_fig}
    \end{subfigure}
    \begin{subfigure}[t]{0.24\textwidth}
    \includegraphics[width=\textwidth]{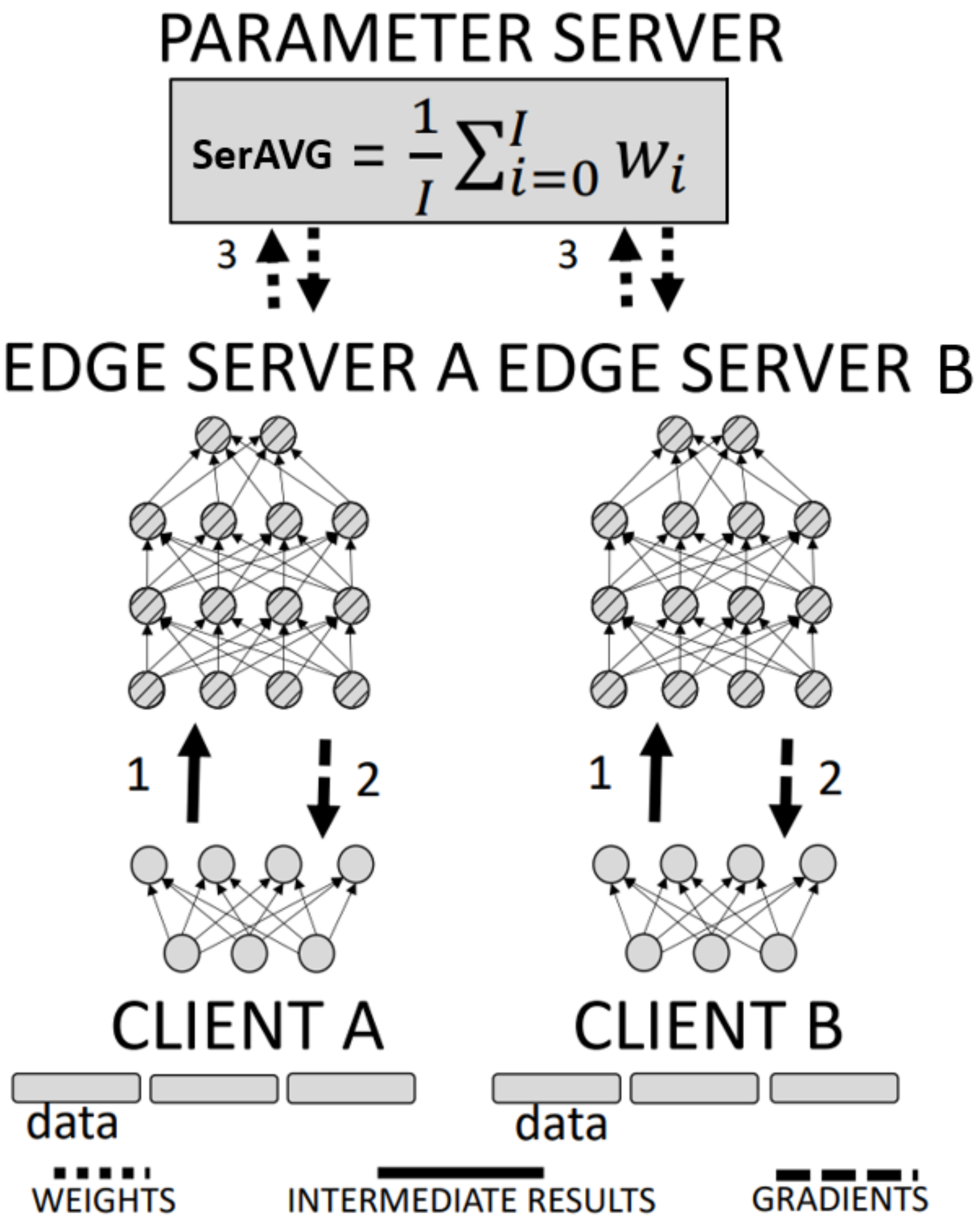}
    \caption{Federated Split Learning}
    \label{FSL_arch_fig}
    \end{subfigure}
    \caption{Distributed NN Training Architectures:
    \textit{(a)~Federated Learning:} The NN is in the client. The parameter server calculates the average weights among clients and overrides the local weights.
    \textit{(b)~Split Learning:} 
    \zz{The server partition sequentially trains with each of the clients. Client weights are shared with the next training client.}  
    \textit{(c)~Parallel Split Learning:} \zz{The server trains clients' output in batches in parallel, but the client's weights are kept private.}
    \textit{(d)~Federated Split Learning:} \zz{Multiple Edge Server and Client pairs train simultaneously. The Edge Servers' weights are averaged by a Parameter Server. The clients weights are kept private. }
    }
    \label{distributed_arch}
\end{figure*}

The FSL model is characterized by multiple edge client -- server pairs.
Such pairs train their copy of the NN simultaneously, providing the parallelism of federated learning, while the client-server separation brings the advantages of the split learning.
Each computing pair partitions the NN.
After some pairs have completed some training epochs, the server NN weights are averaged in a central cloud server, as in classical FL algorithms~\cite{McMahan2017CommunicationEfficientLO}. We call this central cloud node the ``Parameter Server node" in Figure~\ref{FSL_arch_fig}.

There are other techniques, i.e., Parallel Split Learning (PSL)~\cite{parallelSplit}, and Federated Reconstruction (FRC)~\cite{federatedreconstruction}, close to our proposal. 
\zztbd{But they have some disadvantages comparing to FSL.}
While the PSL architecture has only one cloud server node, FSL allows parallel server computations.
FSL and FRC have similar privacy levels (Sec~\ref{sec:attack}), but our evaluations shows that FRC doubles the training time compared to FSL, considering no dominant transmission delays.

We evaluate the benefits of our FSL architecture by testing it with different NN models and tasks: from image classification to an Internet traffic classification~\cite{traffic-classification-survey}.
Aside from testing \zztbd{training performance of} our proposed hybrid federated-split architecture, we evaluated the privacy-performance tradeoff of FSL and SL, and give ideas on how to enhance the privacy guarantee of these schemes using \zztbd{Client-based Privacy Approach (CPA) and} novel neural network partitioning approaches.
Furthermore, we realized that \zztbd{certain ways of partitioning NN}
could reduce transmission delay and enhance privacy together.
Our experiment results show that 
by combining different privacy approaches and NN partitioning methods, our FSL may achieve  both high efficiency with respect to training time, privacy guarantees and accuracy.


The rest of the paper is organized as follows:
in Section~\ref{background-and-related}, we introduce the background and the related works to discuss FSL.
Then, we describe the proposed FSL system in Section~\ref{POFSL}.
In Section~\ref{PAFSL} we discuss the Client-based Privacy Approach (CPA) \zztbd{and motivations to partition NN at edge that can be applied in any split learning based architecture.}
Our evaluation with training and inference metrics is presented in Section~\ref{expr-POA}, while the privacy evaluation is presented in Section~\ref{expr-PAA}.



\section{Distributed Learning Architectures: \\ Background and Related Work}\label{background-and-related}


\noindent
{\bf Federated Learning}~\cite{FedAvgProof-1,FedAvgProof-2-Free-rider,FedAvgProof-3-FedProx,survey-on-FederatedLearning2022, UAV-FL} is a decentralized machine learning technique that trains neural network models using data sources ``owned" by multiple clients (Figure~\ref{FL_arch_fig}).
A logically centralized parameter server holds the latest neural network model, and orchestrates the sharing of its weights between all clients. 

At the beginning of the training phase of a federated learning process, the parameter server sends the same randomly initialized set of neural network weights to each client.
Each client then trains a local model for multiple epochs using its local dataset.
Until the client models have extracted enough features, that is, a given accuracy threshold is reached, the parameter server keeps retrieving, averaging, and overwriting the weights
(Figure~\ref{FL_arch_fig}
-- steps $1$ to $3$).
Thus, the global model could take advantage of the privately-owned datasets which would not transmit through the network.
Moreover, FL has parameters to specify the frequency for the parameter server to average the weights of a certain group of clients.
In this way, system architects can balance the network traffic and model accuracy.
Thus, FL is considered scalable in terms of the number of clients, as long as
such clients have enough computational power and storage resources to meet the training constraints, or Service Level Objectives (SLOs). 

\noindent
\textbf{Split Learning}\zz{~\cite{splitNN}} is a distributed machine learning technique that is characterized by a computational split of the neural network model into two partitions. Each partition could run on a separate computing node, hence splitting the computational resource demand.
To train a NN model with split learning, the NN must first be partitioned \zz{on different nodes}.
Then, the forward propagation phase starts.
When the end of the NN partition at the first node is reached, the outputs of the last layer of activation functions are sent to the server node  --- step 1 (Figure~\ref{SL_arch_fig}).
Then those outputs are used as inputs to the second NN partition and continue the forward propagation.
After calculating the loss, the backward propagation in the server node is started.
Once the backward propagation reaches the input layer of the server's partition, 
the gradients of the inputs are sent to the last activation in the client to finish the backward propagation for client NN --- step 2. 
Finally, the complete NN weights are updated with the gradients.
However, the server's NN partition can also pair with other clients' NN partition with the same client NN structure.
First, the trained weights on the last client are moved to the new client --- step 3. 
Then the new server and client pair trains as mentioned above --- steps 4 and 5. 

\begin{figure}
    \centering
    \includegraphics[width=0.3\textwidth]{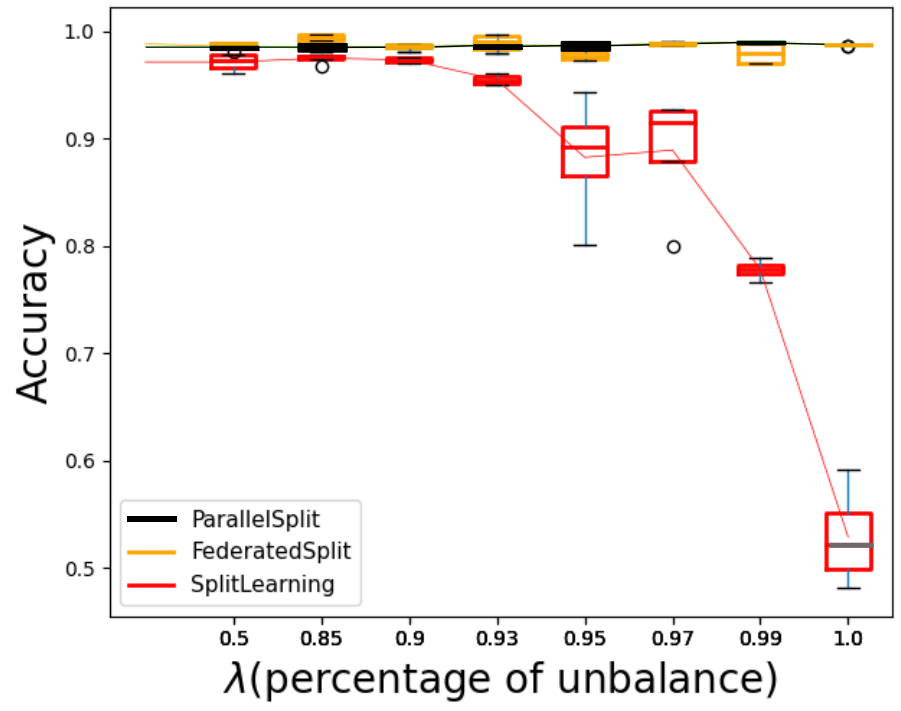}
    \caption{Accuracy degradation when training with Non-IID data.}
    \label{fig:unbalance}
    \vspace{-4mm}
\end{figure}

The SL algorithm preserves data privacy but suffers from a long convergence time with Non-Independent and Identically Distributed (non-IID) data sources (Figure~\ref{fig:unbalance}), and large intermediate data to transmit. Moreover, training with more than one client is sequential, hence poorly scalable. 
%
Some solutions attempted to mitigate the transmission delay of such intermediate data. For example, in BottleNet~\cite{bottlenet} and BBNet~\cite{BBNet}, the authors aim at compressing the intermediate data with a particular NN design.
In Early-Exit~\cite{early-exit} instead the idea is to add classifiers at the early layers to avoid computing the complete model.
Previous work~\cite{sl-distill} also used knowledge distillation to reduce the complexity of client models and the data to transmit.

In this paper, we consider hybrid split-federated learning systems, to combine their benefits while minimizing their drawbacks.  For example, we show that hybrid Federated Split Learning and Parallel Split Learning can converge with non-IID sources faster than Split Learning. 



\noindent
{\bf Combining Split and Federated Learning.}
Other researchers have proposed combining the advantages of split and federated learning.
\zz{
In Parallel Split Learning (PSL)~\cite{parallelSplit} (Figure~\ref{PSL_arch_fig}), they train the client NN partition on multiple edge nodes in parallel. 
During the Forward Propagation phase, the activation function results from different clients are sent to a single remote server. 
Such server then backward propagates the gradient from the loss function to the clients.
}

\vspace{-4mm}
\noindent
\textbf{Model specific approaches.}
FedSL~\cite{FedSL} is one example of Recurrent Neural Networks (RNN).
The idea is to unroll the RNN's feedback loop and split the recurrent NN partition to different nodes with the sequence data segments.
After each epoch, devices average and overwrite their weights.
This approach combines the benefit of SL and FL in the RNN training setting without introducing much overhead. 

\noindent
\zz{\textbf{FL Extensions.}}
Some works extend the FL model instead of designing a hybrid architecture.
The authors of Federated Construction (FRC)~\cite{federatedreconstruction}, prioritize user data privacy trading off the efficiency of the training process.
Their model is partitioned into \textit{global} and \textit{local} shards and both deployed in each edge device, and the two partitions are trained alternately.
A parameter server then sends the global shard and retrieves the corresponding updated weights.
This design makes the training stateless and, consequently, highly scalable for storage.

\noindent
{\bf Source Data Privacy.}
\label{RW:SDP}
One of the advantages of these edge training systems is the high perceived level of source data privacy: 
user's data doesn't leave the edge device.
However, the adversaries can also learn the source data from the NN weights. 
All systems mentioned except for FRC \zz{and PSL} cannot maintain the privacy for NN weights~\cite{model-inversion-attack-1,model-inversion-attack-2}. 
They maintain this privacy since they do not share the complete NN weights or the activation results.

%
\zztbd{Another threat discussed in NoPeek~\cite{nopeekarticle} targets the results of the activation function, which can be used to reconstruct the source data with an autoencoder NN.}
\zztbd{Certain edge intelligence systems that partition Neural Networks and transmit intermediate data in between partitions are vulnerable to this threat.}
Possible mitigation uses the Distance Correlation (DC) loss~\cite{nopeekarticle} added to the original loss function to measure the difference between the source and the intermediate data.
This approach maximizes DC loss and accuracy while updating the NN's weights by solving a Multi-Objective Optimization Problem. 


\noindent
{\bf Client-Based Privacy Approach (CPA).}
NoPeek's loss function improves the resilience in reconstructing private source data from the intermediate data.
However, sharing the loss and constructing the distributed gradient graphs are either unsafe or need an extra management system.
To overcome this limitation, we propose \zztbd{an extension from NoPeek, namely Client-based Privacy Approach.}
The idea is to add noise \zztbd{with user specified methods} to the results of activation functions in clients, so there would be less features to be used by the autoencoder NN \zztbd{(Section~\ref{CPA})}.


\noindent
\zz{{\bf Privacy Approaches.} 
We will mainly compare DC and Differential Privacy (DP-SGD) under CPA framework.
DP-SGD is a widely used lightweight algorithmic approach for data privacy~\cite{DLwithDP,DPDNN}. 
The idea is to add a Gaussian noise to the gradients during a training phase of a NN.
Thus the client's output, generated by the updated weights, will confuse the adversary.
Comparing to Homomorphic Encryption (HE) or Secure Multi-Party Computation (SMPC), we consider it fits better in our client and edge server setup, while the privacy level of encryption has been well discussed \cite{DPDNN}. 
It is not cost efficient at edge to consume a lot of battery for encryption and decryption steps.
There are also methods\zztbd{, i.e., model compression(\cite{BBNet,CLIO,sl-distill}),} can potentially enhance privacy guarantee and consume power lower than HE but higher than DP-SGD .
But comparing existing privacy methods is not the focus of this project and we consider it as a future work.
We mainly want to show the proposed CPA is a general approach and it provides high attack resilience with different methods including DC and DP-SGD.
Thus, in this paper, we evaluate the traditional DP-SGD applied to the complete model 
 and compare it with DP-SGD only applied to the clients using the CPA framework (Section~\ref{eval-CPA-DP}).
}


\section{Privacy-Oblivious FSL}\label{POFSL}
In this section, we detail our proposed Federated Split Learning (FSL) architecture, that we originally proposed in~\cite{FSL}. 
FSL is a hybrid approach that combines the advantages of SL and FL. 
\zz{It avoids sending users' source data or sharing the complete NN parameters through the network while being scalable.}

Our FSL architecture shown in Fig.~\ref{FSL_arch_fig} has three types of entities: $(i)$ edge servers, $(ii)$ clients, and $(iii)$ parameter servers.
To train a NN with FSL, we first setup a\sai{n} authentication protocol~\cite{auth-protocol-1, auth-protocol-2} among the entities to pair 
\zz{each client with one edge server, and edge servers with a parameter server.}
After pairs are found, the communications are sent without encryption.
Then, in each client and server pair, we partition the complete NN into the client's partition and (edge) server's partition.

\zz{FSL has three training steps. In step 1, the client forward propagates with source data and transmits the intermediate data to the edge server. The server then finishes the propagation and calculates the loss.
In step 2, the server backward propagates to client source data.
In step 3, after epochs, the {\it parameter server} averages the weights in the edge servers. 
}

\zz{Consequently, FSL will have multiple advantages compared to the other approaches discussed. 
FSL clients will have a lower resource demand compared to FL since it will have \sai{fewer} NN layers to train. 
\zztbd{Thus FSL is a more practical scheme for edge intelligence application.
Also, w}hile FSL only averages the weights in the edge servers, FedAVG averages the complete weights.
\zztbd{Therefore} FSL avoids potential vulnerability to model inversion attacks~\cite{model-inversion-attack-1, model-inversion-attack-2}.
\zztbd{Moreover, }FSL provides better scalability than SL, since client and server pairs can train independently.}
Compared with the FRC~\cite{federatedreconstruction} architecture, we note that the latter is inefficient in training time. It updates the parameters of one partition on each forward and backward propagation execution and runs multiple times to update the full model.
Compared with \zz{Parallel Split Learning~\cite{parallelSplit}}, we found four \zz{potential suboptimalities.}
First, since the edge devices have to synchronize with the central server,
clients may have to wait until the server has finished with processing all 
the results of the activation functions in its queue.
In the worst case, assuming that all  activation function results from $n$ clients arrive at the same time, the lower bound of waiting time for each client to continue on with backward propagation is $O(n)$. We hence conclude that PSL is not as scalable as our FSL.
Second, we also observe that the PSL server would temporarily store multiple batches of the results of activation functions, so it needs a sophisticated logging and compaction storage system to recover from failures.
Consequently, PSL is less robust than our FSL, since 
FSL has to maintain fewer states in each isolated pair.
Third, the PSL design may suffer from resiliency problems.
\zztbd{Meanwhile, in FSL, failure in one pair won't prevent other pairs from training or inference.}
Fo\sai{u}rth, in PSL, since all intermediate data will be transmitted and processed by the single server, bandwidth and computation resource at the server node may get congested. 
\zz{Our findings are presented in Section~\ref{EVAL-Sec}.}

\vspace{-2mm}
\section{Privacy-Aware FSL}\label{PAFSL}

We have discussed the efficiency and fault-tolerance properties of FSL. In this section, we consider instead the privacy-preserving properties of different architectures and propose our privacy-aware FSL.
In particular, we discuss how to complement general split learning based architectures to mitigate the problem of sharing the output values of NN activation functions or weights over an honest but curious network.
We first give a formal definition of our privacy attacker model, and then we discuss how a Client-based Privacy Approach and \zztbd{certain ways of partitioning Neural Network would help avoid such attack.}

\vspace{-3mm}
\subsection{Privacy Attacker Model and Assumptions}\label{sec:attack} 

We assume an attacker can capture the \textit{Intermediate Data} (the results of last layer activation functions transmitted from client to edge server) in plaintext. Moreover, we assume that \zztbd{attacker knows the client NN architecture.}
Consequently, the adversary can implement an AutoEncoder~\cite{autoencoder} NN to reproduce the source data fed into the client model.
\zztbd{Moreover, to train the autoencoder NN, we assume that some datasets with features similar to the client source data are accessible to the adversary.}

\vspace{-7mm}
\zz{
\subsection{Attack Resilience}\label{resilience_tau}
Given the attacker model, in order to compare the level of privacy guarantee among different privacy approaches, we define an \textit{Attack Resilience} metric ($\tau$) as:
\begin{equation}\label{tau}
\tau = 1-\frac{\|correct\|}{\|reconstructed\|}
\end{equation}
It measures the misclassification rate. 
$\|correct\|$ counts the number of images, reproduced by the attacker, which can be correctly classified by a trained classifier (Section \ref{sec:PA-methodology}).
And $\|reconstructed\|$ is the total number of reproduced images.
}

\vspace{-2mm}
\subsection{Client-Based Privacy Approach in Distributed Setting via Distance Correlation (CPA-DC)}\label{CPA}

Motivated by the NoPeek approach~\cite{nopeekarticle}, where  weights are  updated based on the sum of two loss functions, i.e., Cross-Entropy, and  Distance Correlation (DC), we adopted the idea of using two loss functions into an alternately scheduling mechanism with two rounds:
the \textit{regular round},  minimizes the cross-entropy loss function in the (edge) server. The \textit{Distance Correlation round}, maximizes the DC loss function in the client.
The idea is that we add noise that make source and intermediate data different to client's weights.
%
We define this alternating behavior with loss functions in Equation~\ref{CPADC_loss_function}:


\vspace{-2mm}
\begin{equation}\label{CPADC_loss_function}
L=
\begin{cases}
Loss(g(f(x)), label) & if \hspace{0.2cm}  e~$mod$~F~$== 0$\\
m \cdot DC(x, f(x)) & \text{otherwise,}
\end{cases}
\end{equation}
where $L$ is the measured loss value, $e$ represents the epoch index, $F$ represents the $DC\_Frequency$, $Loss(\cdot)$ is the loss function used to measure mis-classifications, $DC$ is the distance correlation loss, $m$ is the \textit{loss multiplier}, $f$ is the client NN, $g$ is the server NN, $x$ is the source data, and $label$ is the labels in the source data.
\zz{Notice that $F$ controls the alternation frequency and $m$ adds a weight to the DC loss.}
The alternating loss function is a policy. Our Client-Based Privacy Approach (CPA) can also work with other loss functions or methods adding random noise to the regular round.
In the evaluation, we explored several methods to embed the noise.
In particular, we explored the  trade-off between training time and the highest attack resilience.
\vspace{-2mm}
\subsection{How many layers do we assign\\ to each neural network partition?} \label{NPS}

In this section, we discuss the problem of selecting how many layers need to be assigned for each NN partition\zztbd{, i.e., client NN depth}.
This tradeoff will tune training time \zztbd{(processing and transmission delay)}, privacy, and accuracy.
Capturing the tradeoff between all these metrics is challenging.
To illustrate,
consider the tradeoff between processing delay and transmission delay.
\zztbd{
The size difference among output layers in different partitions can be large, so a few partitioning policies may lead to significant transmission overhead, increasing training time and hence diminishing the gain of the hybrid FSL compared to the original Federated Learning architecture.
In VGG-16\cite{VGG16}, the output size of the first convolutional layer is two times the size of the second convolutional layer.
Thus, a system with a model cut after the second convolutional layer can tradeoff the extra processing delay at low-capacity clients while yielding a lower transmission delay. We evaluate this effect in Section~\ref{latency-eval}.}

\zztbd{
Analytically, this effect is captured by solving
Problem~\ref{NPS_opt_problem_single_cut},
where
$\alpha$ and $\beta$ are developer-specified parameters that represent positive weights for the transmission delay of intermediate data (I) and computation delay (C), the parameter $d$ represents the depth of the client neural network, and finally, $b$ represents the bandwidth, which is measured periodically.
To efficiently solve this problem,
we follow the approach in~\cite{Neurosurgeon},
where
we build two regression models to predict the delays I and C, given the available bandwidth by profiling the model, i.e., computing the output size and processing time for each layer in the client NN, instead of training the full model.
}
\pstwo{
\begin{equation}\label{NPS_opt_problem_single_cut}
\begin{split}
\min_{d}(\alpha I(d \mid b) + \beta C(d))
\end{split}
\end{equation}
}

\vspace{-5mm}
\zztbd{
The solution of Problem~\ref{NPS_opt_problem_single_cut} is optimal with respect to delays, however, it can be sub-optimal with respect to privacy and accuracy.
As Section \ref{latency-eval} shows, the client processing and transmission delay of FSL reach the minimum when the client NN depth is between 7 and 16.
In Fig.~\ref{accu&resi_CIFAR10_VGG16_oblivious}, instead, the client NN needs more than 16 layers to be above $90\%$ attack resiliency.
Thus, we conclude that an optimal NN partitioning decision should balance different objectives and constraints, including transmission delay, processing time, privacy, and accuracy, as shown in our Problem formulation~\ref{NPS_full_opt_problem_single_cut}.
In such a problem, $W$ represents the model weight vector, $(I', C', A, R)$ is the tuple representing the observations for transmission delay, computation delay, accuracy, and resilience, $(\gamma, \kappa)$ are new user-specified positive weights, and $d$ is the client NN depth.
%
%
\begin{equation}\label{NPS_full_opt_problem_single_cut}
\begin{aligned}
\max_{W,d}\quad &( -\alpha I'(W, d \mid b) - \beta C'(W, d) \\
                &+ \gamma A(W, d)+ \kappa R(W, d))
\end{aligned}
\end{equation}
%
}

\zztbd{
To solve Problem~\ref{NPS_full_opt_problem_single_cut},
we have to train $W$ for each $d$ until convergence and then find the best $d$.
This brute force method is inefficient.
A more efficient approach would rely on predicting the delays, accuracy, and privacy without the full training of the model. 
Extending the approach in~\cite{Neurosurgeon} to go beyond profiling delays,
is challenging.
This is because the accuracy and attack resilience for each client and edge server pair is harder to profile and predict. 
Specifically, their profiling depends on the weights trained on other pairs,
the distribution of source data among clients, 
number of clients, number of layers to average in SerAVG, and 
training epochs (Sec.~\ref{accu-eval} and \ref{expr-PAA}).
Another work can predict the model accuracy\cite{PredictAccu}, 
but it is based on the already trained model.
Therefore, for our FSL architecture with SerAVG, a prediction method for partitioning remains an open question for future work.
In this paper, we experimentally demonstrate the best model partitioning that balances requirements on training time, accuracy, and privacy.
}

\section{Evaluation Results}\label{EVAL-Sec}
\zz{In this section, we describe the evaluation results related to our Privacy-Oblivious FSL (POFSL) and Privacy-Aware FSL (PAFSL) architectures with our privacy-aware approaches (CPA in Section~\ref{CPA} and Neural Network Partitioning in Section~\ref{NPS}).
Our evaluation demonstrates the advantages of FSL over PSL and FRC in terms of training time, memory 
usage, and convergence rate.
Moreover, we also show that our privacy-aware approaches can prevent the reconstruction of source images from intermediate data in the Split Learning-based systems.
We first discuss our experimental setup, then present our evaluation results of POFSL and PAFSL in Sections~\ref{expr-POA} and \ref{expr-PAA}.
}

\subsection{Experimental Setup}\label{experimental-setup}
\zztbd{This experiment set studies the convergence for POFSL and the privacy guarantee of PAFSL across different hardware and applications with different NNs and datasets.
}

\zz{
For the hardware, we used two types of nodes on Chameleon Cloud~\cite{articleChameleon}.
One has an RTX6000 GPU, two Intel Xeon Gold $6126$ CPUs
and $187$ GB memory.
The other one has four NVIDIA V100 GPUs, two Intel Xeon Gold 6230 CPUs 
and 128 GB of memory.
We emulated the computer network among our distributed learning entities on the localhost interface on a physical machine, and each experiment was set to use a single GPU. So that we can ignore the network bandwidth bottlenecks.
}

\begin{figure*}[t]
\captionsetup[subfigure]{justification=centering}
    \centering
    \begin{subfigure}[t]{0.19\textwidth}
    \centering
    \includegraphics[width=\textwidth]{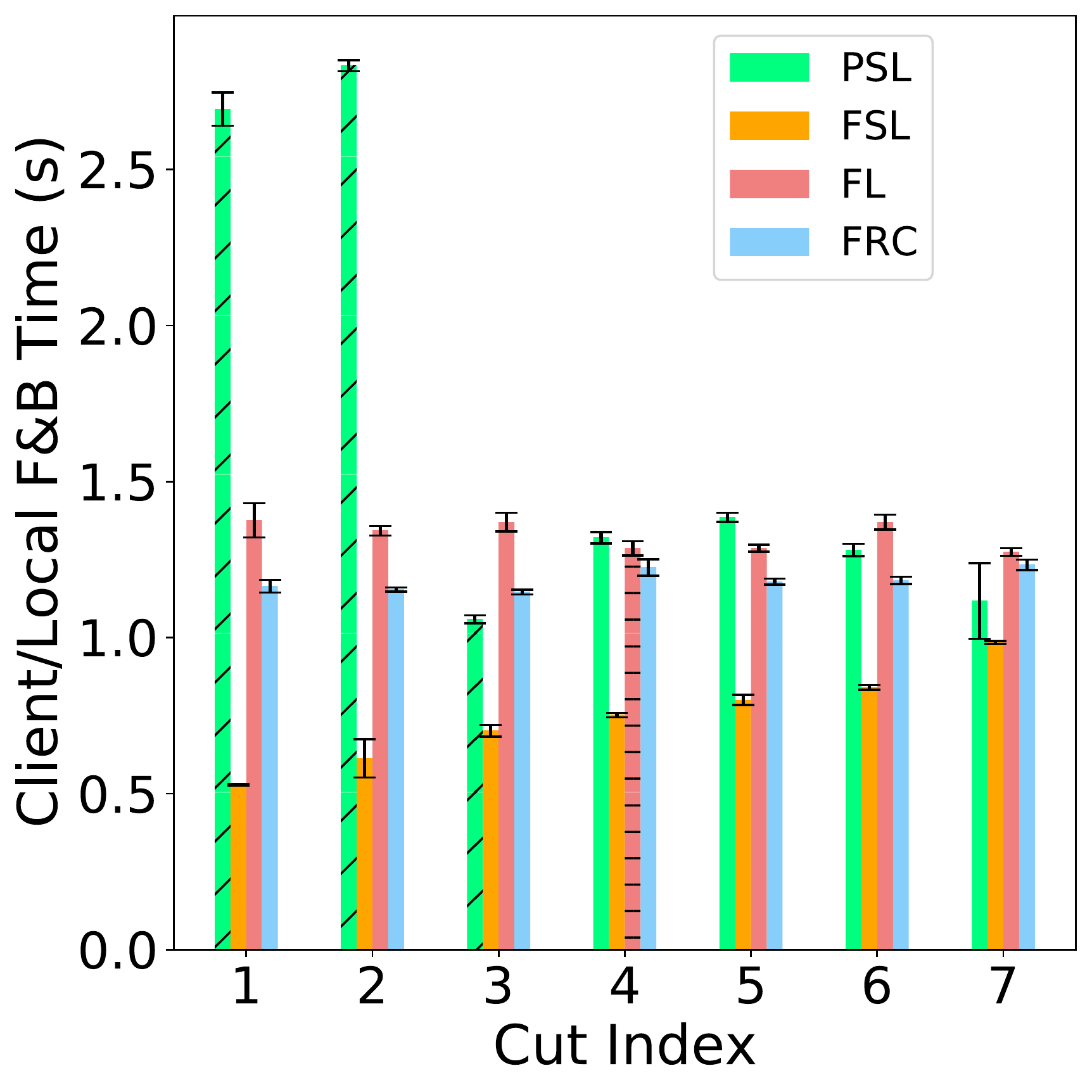}
    \caption{LENET+MNIST \\Client Measurements}
    \label{LeNet+MNIST_time_CUTS_Client}
    \end{subfigure}
    \begin{subfigure}[t]{0.19\textwidth}
    \includegraphics[width=\textwidth]{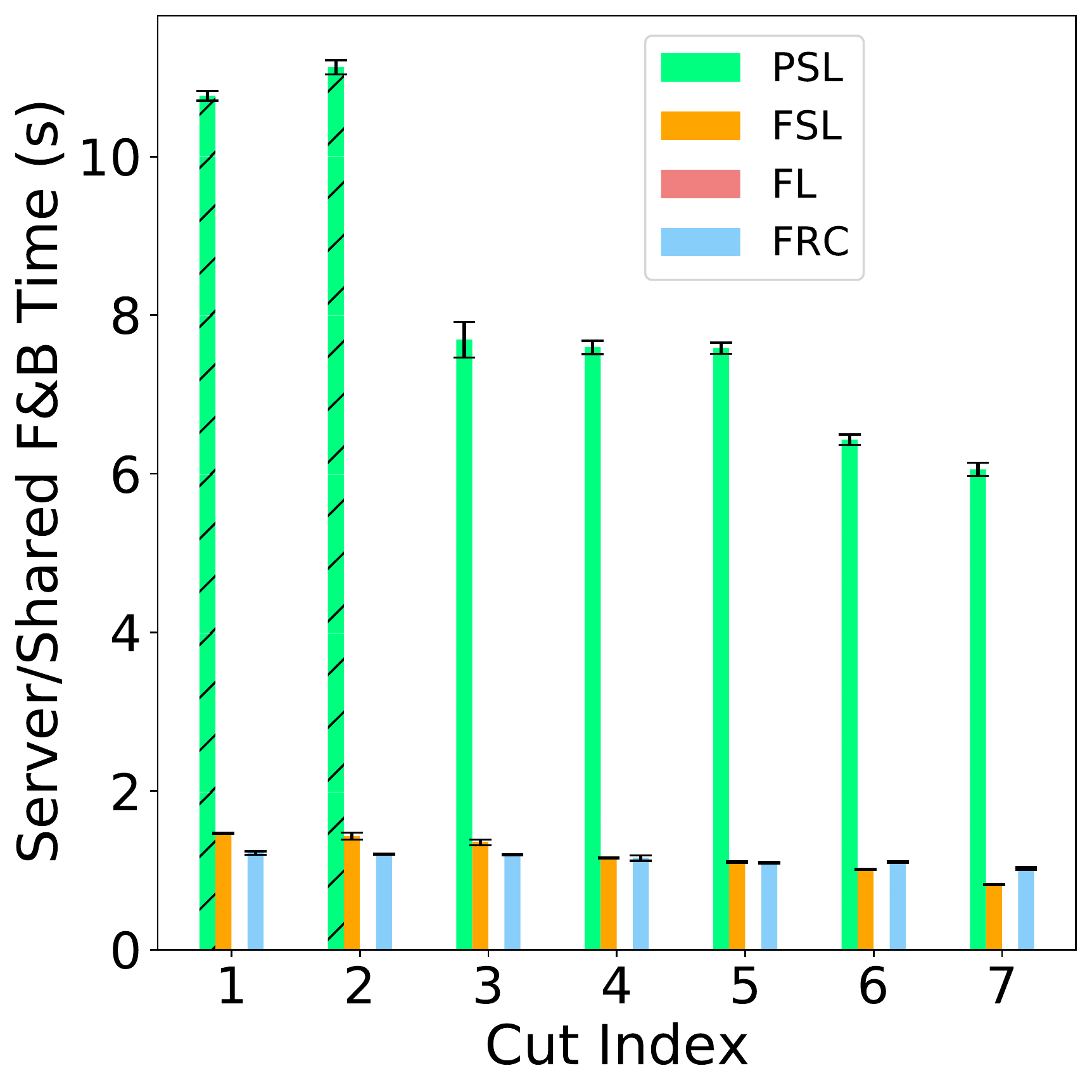}
    \caption{LENET+MNIST \\Server Measurements}
    \label{LeNet+MNIST_time_CUTS_Server}
    \end{subfigure}
    \begin{subfigure}[t]{0.19\textwidth}
    \includegraphics[width=\textwidth]{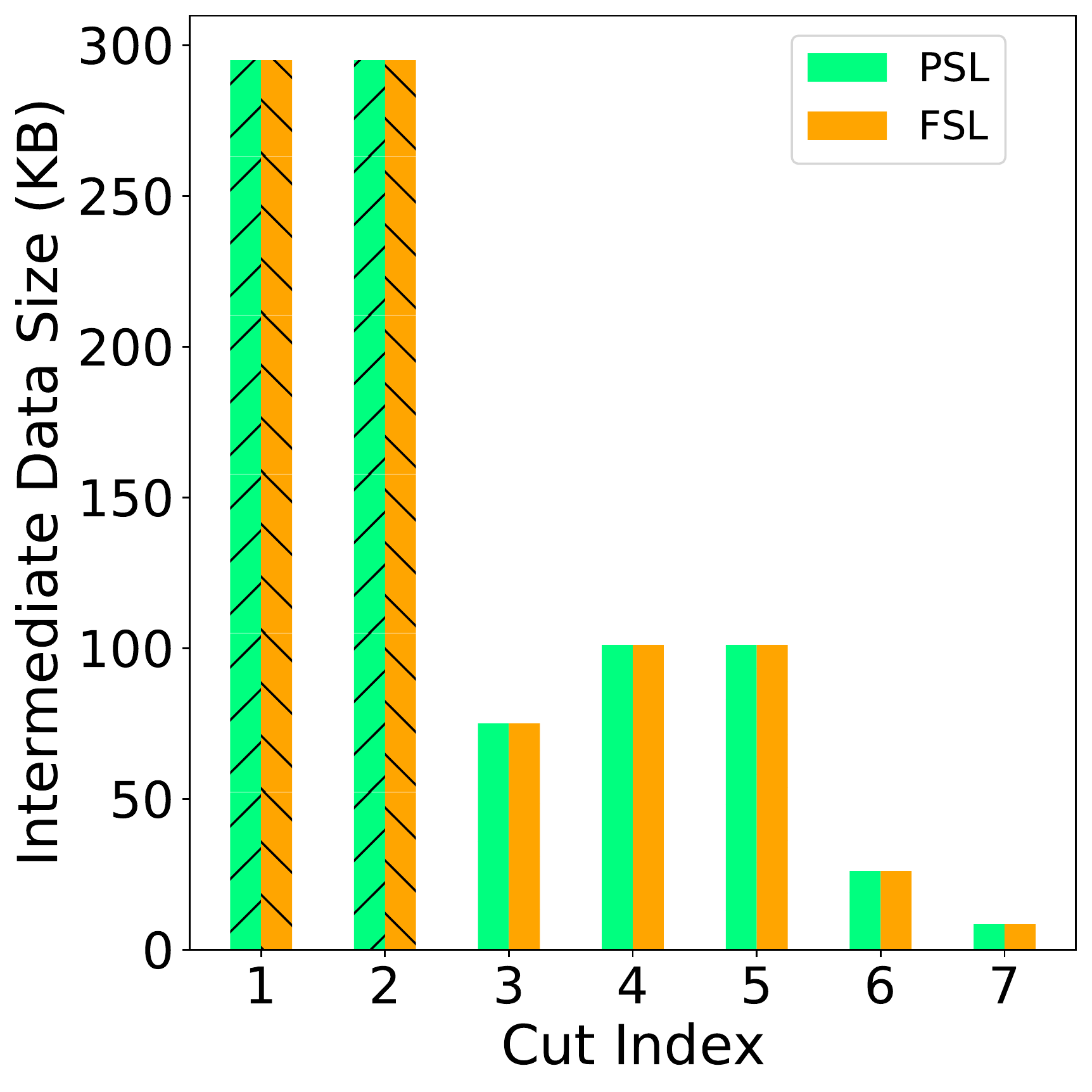}
    \caption{LENET+MNIST \\Intermediate Data}
    \label{LeNet+MNIST_time_CUTS_Interm}
    \end{subfigure}
    \begin{subfigure}[t]{0.19\textwidth}
    \includegraphics[width=\textwidth]{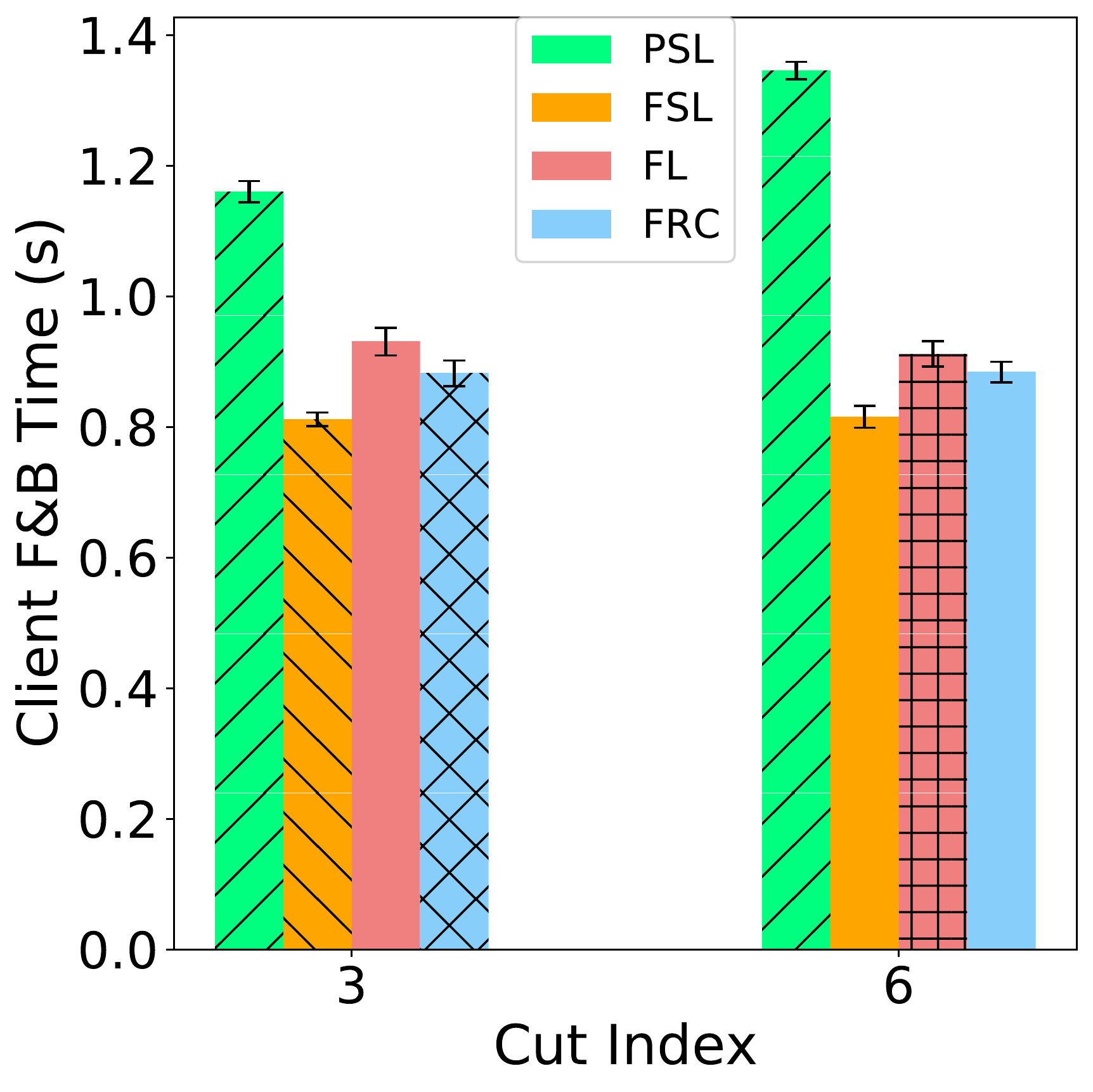}
    \caption{VPN Workload \\Client Measurements}
    \label{VPN_Client}
    \end{subfigure}
    \begin{subfigure}[t]{0.19\textwidth}
    \includegraphics[width=\textwidth]{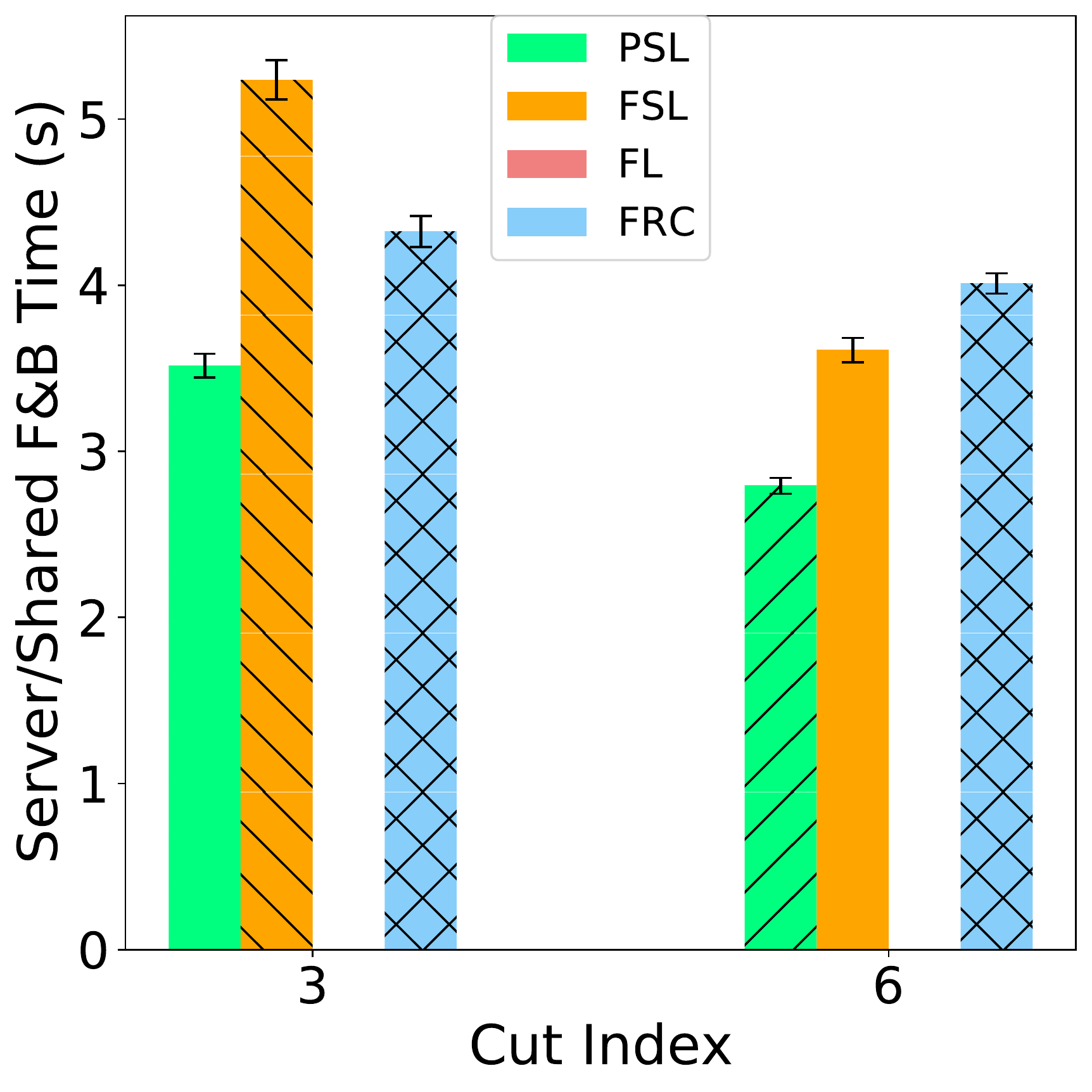}
    \caption{VPN Workload \\Server Measurements}
    \label{VPN_Server}
    \end{subfigure}
    \caption{\textit{LeNet+MNIST}: (a)~Client Time, (b)~Server Time, (c)~Intermediate data size. 
    Observations: 1) Intermediate data size is correlated with the times taken by the PSL architecture while having \zz{little} correlation under FSL;
    2) FRC has almost twice the overall training time as FL; 
    3) Plots are obtained by averaging 20 clients' results. Intermediate data size is under batch size of 16 and each image was resized to (1,32,32). 
    \textit{VPN Workload}: (d)~Client Time (e)~Server Time. Similar considerations are valid for the VPN dataset. Tested with 5 clients with a one dimentional NN with input size of (1,784). Still, FSL has the shortest Client F\&B time compared to the other settings.
    }
    \label{LeNet+MNIST_time_CUTS}
    \label{VPN}
    \label{Training-Time-Evaluation-Ph1}
\end{figure*}
\begin{figure}[t]
\captionsetup[subfigure]{justification=centering}
    \centering
    \begin{subfigure}[t]{0.22\textwidth}
    \includegraphics[width=\textwidth]{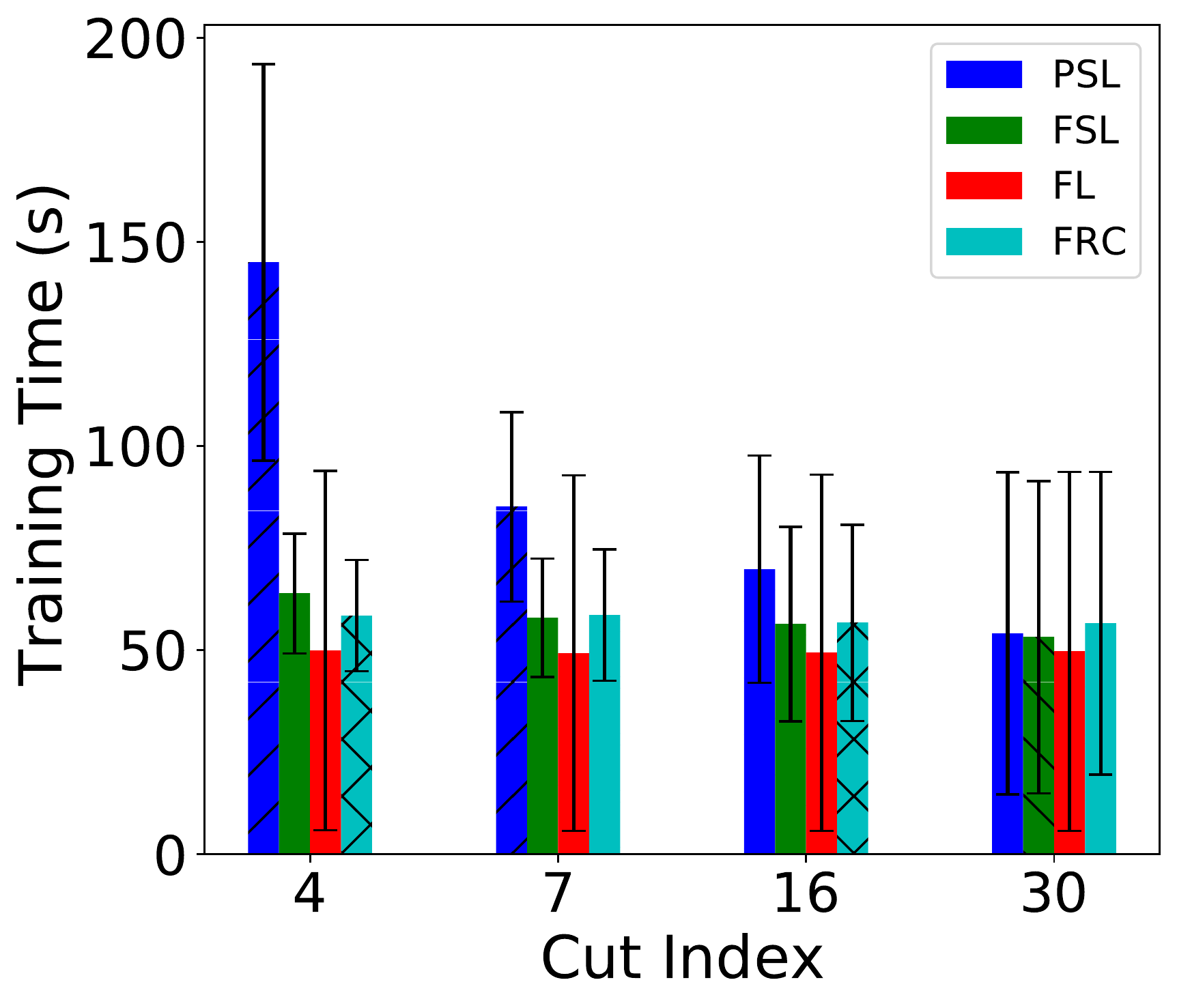}
    \caption{Pair Time}
    \label{VGG16+Cifar-10-breakdown-pair-time}
    \end{subfigure}
    \begin{subfigure}[t]{0.22\textwidth}
    \includegraphics[width=\textwidth]{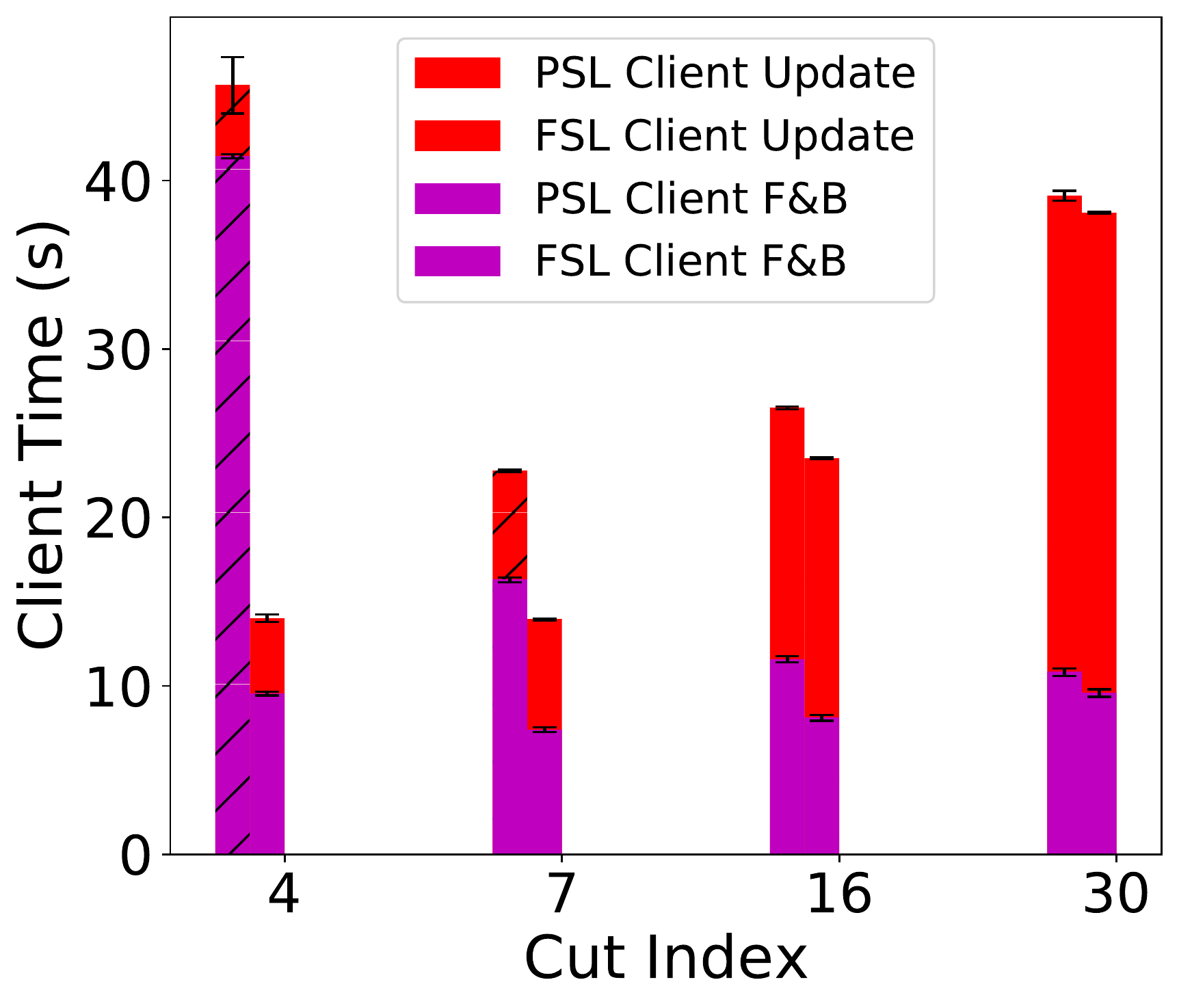}
    \caption{Client Time}
    \label{VGG16+Cifar-10-breakdown-client-time}
    \end{subfigure}
    \begin{subfigure}[t]{0.22\textwidth}
    \includegraphics[width=\textwidth]{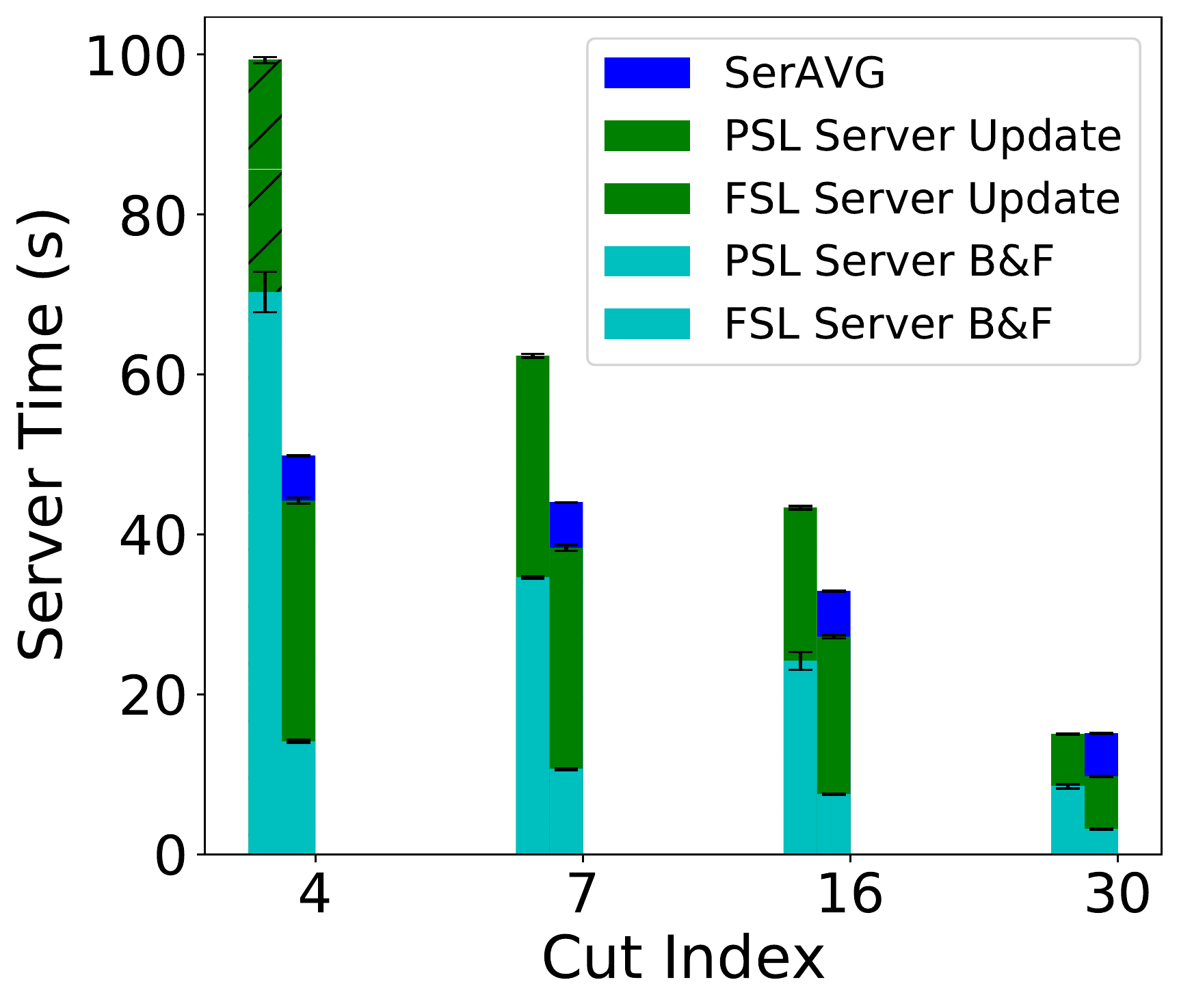}
    \caption{Server Time}
    \label{VGG16+Cifar-10-breakdown-server-time}
    \end{subfigure}
    \hspace{3mm}
    \begin{subfigure}[t]{0.19\textwidth}
    \includegraphics[width=\textwidth]{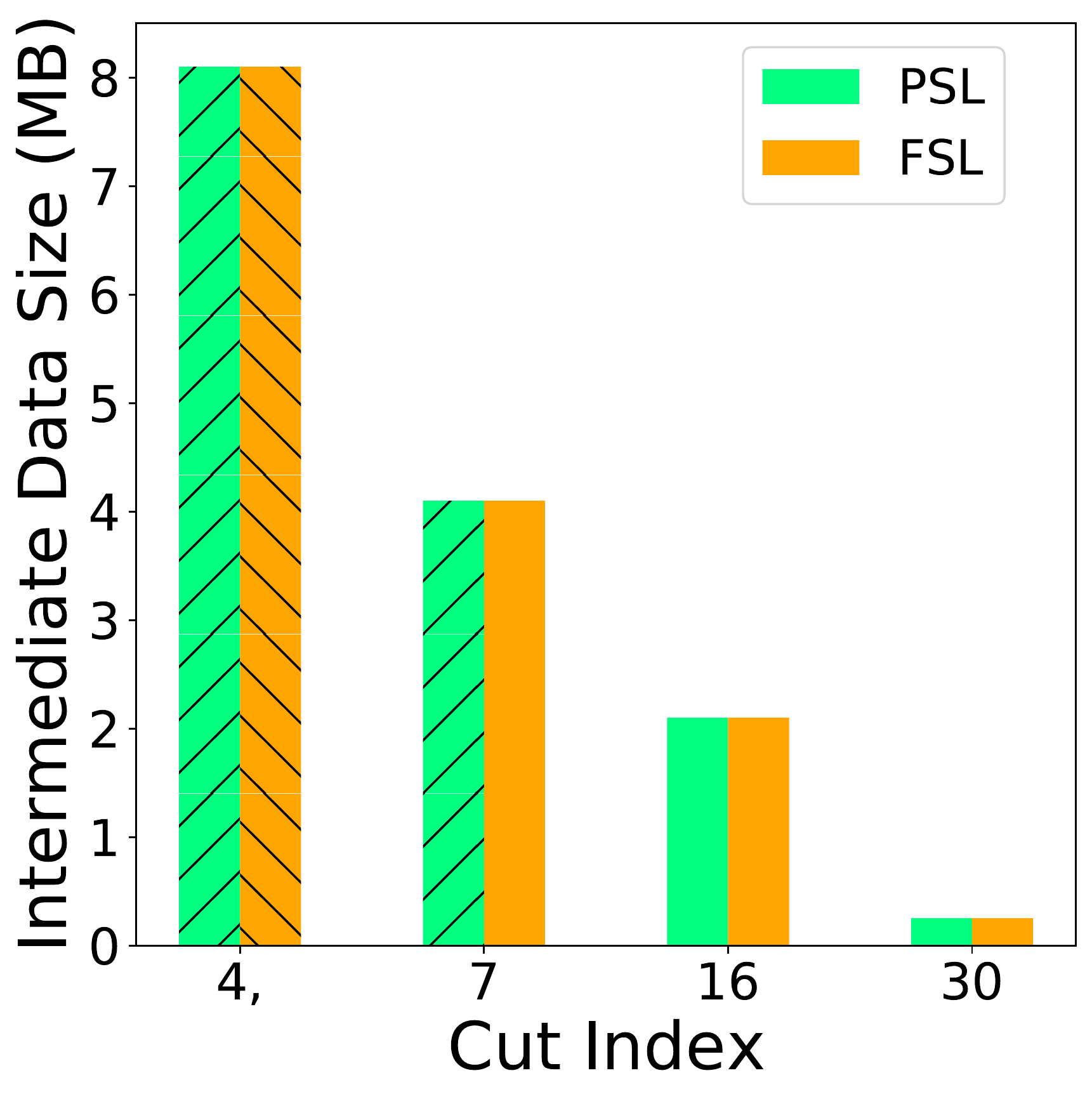}
    \caption{Intermediate Data Size}
    \label{VGG16+Cifar-10-breakdown-interm-size}
    \end{subfigure}
    \caption{VGG + CiFar10: Plots show the effect of Cut Index over 
    (a)~averaged overall training time, (b)~time spent in clients (5 clients average), (c)~time spent in server, (d)~intermediate data size. 
    The transmission delay caused by intermediate data size can dominate the training time (occurring during F\&B propagation). 
    }{}
    \label{VGG16+Cifar-10-breakdown}
    \label{Training-Time-Evaluation-Ph2}
    \vspace{-2mm}
\end{figure}

\zz{
\zztbd{For the applications, we considered three classification tasks and implemented with PyTorch~\cite{pytorch}.}
Then the distributed communication among entities of the systems was handled by PySyft~\cite{PySyftWhitePaper} and  PyGrid~\cite{PyGridRepo} and no encryption is applied on the transmission.
The first application runs a general image classification task with a VGG-16~\cite{VGG16} Convolutional Neural Network (CNN). The model was pre-trained using Imagenet~\cite{ImageNet} and then trained with the CIFAR-10 dataset~\cite{cifar10}. 
We run this task on 5 clients running a NVIDIA V100 GPU.
The second task uses a LENET~\cite{lenet} CNN to recognize handwritten numbers in the MNIST dataset~\cite{lecun-mnisthandwrittendigit-2010} on 20 clients running a NVIDIA RTX6000 GPU. 
The third task classifies traffic, not images. In particular, we decomposed
a one-dimensional-CNN, trained with the ISCX VPN-nonVPN (ISCX) traffic dataset~\cite{traffic-classification}, using 5 clients running on a RTX6000 GPU.
We partitioned the dataset and assign\sai{ed} 
among different clients 
with Independent and Identically Distributed (IID) probabilities and all our plots show $95\%$ confidence intervals, unless otherwise specified. 
Our goal is to verify that FSL can always converge, with different tasks, different NNs, different devices, and different data distributions. We verified the advantages in delay or privacy of FSL over existing solutions.
}

\vspace{-3mm}
\subsection{Evaluation Results for Privacy-Oblivious FSL }\label{experiment-result}\label{expr-POA}
This section illustrates the methodology and draws observations of our experiments. Overall, our evaluations show that POFSL has less overhead and \zztbd{similar} accuracy \zztbd{comparing to existing solutions.}
In particular, we evaluate training time (Sec.~\ref{latency-eval}), memory consumption (Sec.~\ref{mem-eval}), and learner accuracy (Sec.~\ref{accu-eval}).    

\vspace{-2mm}
\subsubsection{Experiment Design}\label{expr-design}
Given the size of the different datasets and number of clients, to reach at least $90\%$ accuracy, the neural networks used for image classification needed 20 epochs. While for the traffic classification model, 80 epochs were used. 

\vspace{-2mm}
\subsubsection{Training Time Evaluation}\label{latency-eval}


To evaluate training time, let us consider the experiment whose results are reported in Figures~\ref{Training-Time-Evaluation-Ph1}~and~\ref{Training-Time-Evaluation-Ph2}. 
The x-axis indicates the {\it Cut Index}, i.e., the index of the last layer running in the client/local part of the NN.
When tested over the MNIST scenario, we can observe from Figure~\ref{LeNet+MNIST_time_CUTS_Client} and \ref{LeNet+MNIST_time_CUTS_Server}
that FSL has the shortest ``Client Forward and Backward Propagation" (Client F\&B) time among all other distributed architectures. 
\zz{The Client F\&B time includes transmission time for gradients and computing both activations and gradients in Client NN. And Server F\&B time includes transmission time for hidden variable from client to server and computation in Server NN. Notice that the weight update time is separately counted by ``Client Update Time`` and ``Server Update Time``.}
Moreover, we note that PSL is more vulnerable than FSL to limited bandwidth across splits.
PSL is consistently the slowest, due to its inefficient server design; the server has to synchronize the intermediate data, and it must process all batches of intermediate data in each training epoch sequentially.

Observing FRC and FL, we see the F\&B times do not change along with the Cut Index (Figures~\ref{LeNet+MNIST_time_CUTS_Client} and \ref{LeNet+MNIST_time_CUTS_Server}).
Note also that FRC is not training time efficient.
It updates its complete model with two almost full forward and backward steps~\cite{federatedreconstruction}.
This can be noted in the same figures: the FRC total F\&B  time for local and shared weights is almost doubled compared to the FL training time.



We were able to obtain similar results comparing the F\&B times on another predicting scenario: the 1D CNN implemented by~\cite{traffic-classification} (Figures~\ref{VPN_Client}~and~\ref{VPN_Server}). 
Due to the limited size of this neural network (with only two convolutional layers), we evaluated the architectures with merely two Cut Indexes: at layer 3 and layer 6 of the NN.  
Even in this experiment, we observe how our FSL still has the shortest Client F\&B time. PSL is the worst performant at each cut, and FL keeps performing better than FRC.  

\noindent
{\bf FSL consistently uses less time in each training epoch than the other analyzed architectures.} {\it We found that PSL perform worse than FSL because of the single-server architecture. PSL has 
similar results when comparing its client F\&B time with FL and FRC.
FRC is not training time efficient. 
Its total F\&B time almost doubles compared to FL.}

When evaluating the training time on the CIFAR-10 scenario, we found a different trend (Fig.~\ref{VGG16+Cifar-10-breakdown-pair-time}): PSL had the longest training time, except for cut index of 30.  
Moreover, FSL did not always perform the best. When most of the layers run within the client, FL has a shorter training time.
\zz{This is because the size of intermediate data changes as the cut moves, and with smaller data to send, the overall training time can be shorter.
Fig.
~\ref{VGG16+Cifar-10-breakdown-client-time} and~\ref{VGG16+Cifar-10-breakdown-server-time}
show the extra F\&B time during training.
And existing works for SL have discussed the similar behavior~\cite{BBNet, bottlenet, sl-distill, early-exit}.}

In particular, we show that FSL 
outperforms PSL as PSL F\&B time is more vulnerable to the intermediate data transmission.
In Figure~\ref{LeNet+MNIST_time_CUTS}, Client F\&B time of FSL keeps decreasing with smaller Cut Index, while that of PSL still increases at 
Cut Index 6, 5 and 3, 2, although the intermediate data in this experiment is much smaller than using VGG-16.
This behavior is caused by the single server bottleneck.
Thus, FSL is more scalable in terms of training time.
Such observation also explains why both client and server F\&B times of PSL are consistently larger than FSL.\\

\vspace{-4mm}
\noindent
{\bf The intermediate data and gradients can cause significant network overhead.} Such overhead,  however is better mitigated by our FSL than PSL.
%
%
%
\zz{Existing work~\cite{BBNet}~\cite{early-exit} for Split Learning, as well as our partitioning strategies (Section~\ref{NPS}) can further mitigate the communication overhead.}
Thus, the additional delay in FSL is not considered a severe bottleneck compared to those systems training at the edge, like FL.

\vspace{-1 mm}
\subsubsection{Memory Consumption Evaluation}\label{mem-eval}


Memory usage of each entity in the edge training and inference systems limits the scope of devices that can join the system.
To compare which system is more flexible to deploy in terms of memory capacity on devices, we show that each entity has calculated memory usage in the FSL, PSL, FL, and FRC systems.

Real-world memory utilization can be highly variable as it depends on several implementation factors, such as libraries used and the Remote Procedure Calls (RPCs) implemented.
However, the size of a model and its activation at each layer are known. 
The following results show that FSL's clients consume less memory compared to FL and FRC, and any of its servers occupy less memory than PSL.



\begin{figure}[t]
    \captionsetup[subfigure]{justification=centering}
    \centering
    \begin{subfigure}[t]{0.22\textwidth}
    \includegraphics[width=\textwidth]{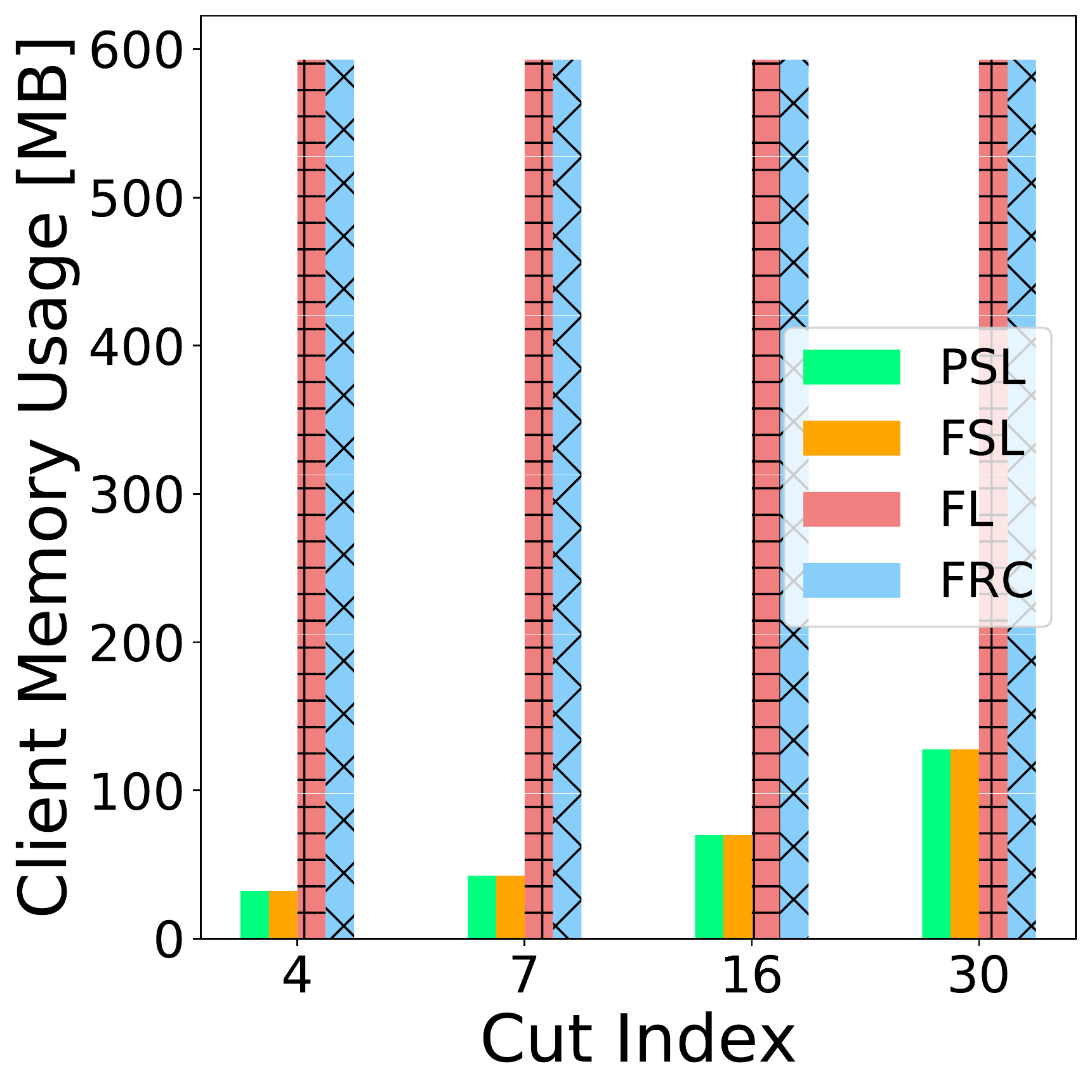}
    \caption{}
    \label{VGG16+Cifar-10-Memory-a}
    \end{subfigure}
    \begin{subfigure}[t]{0.22\textwidth}
    \includegraphics[width=\textwidth]{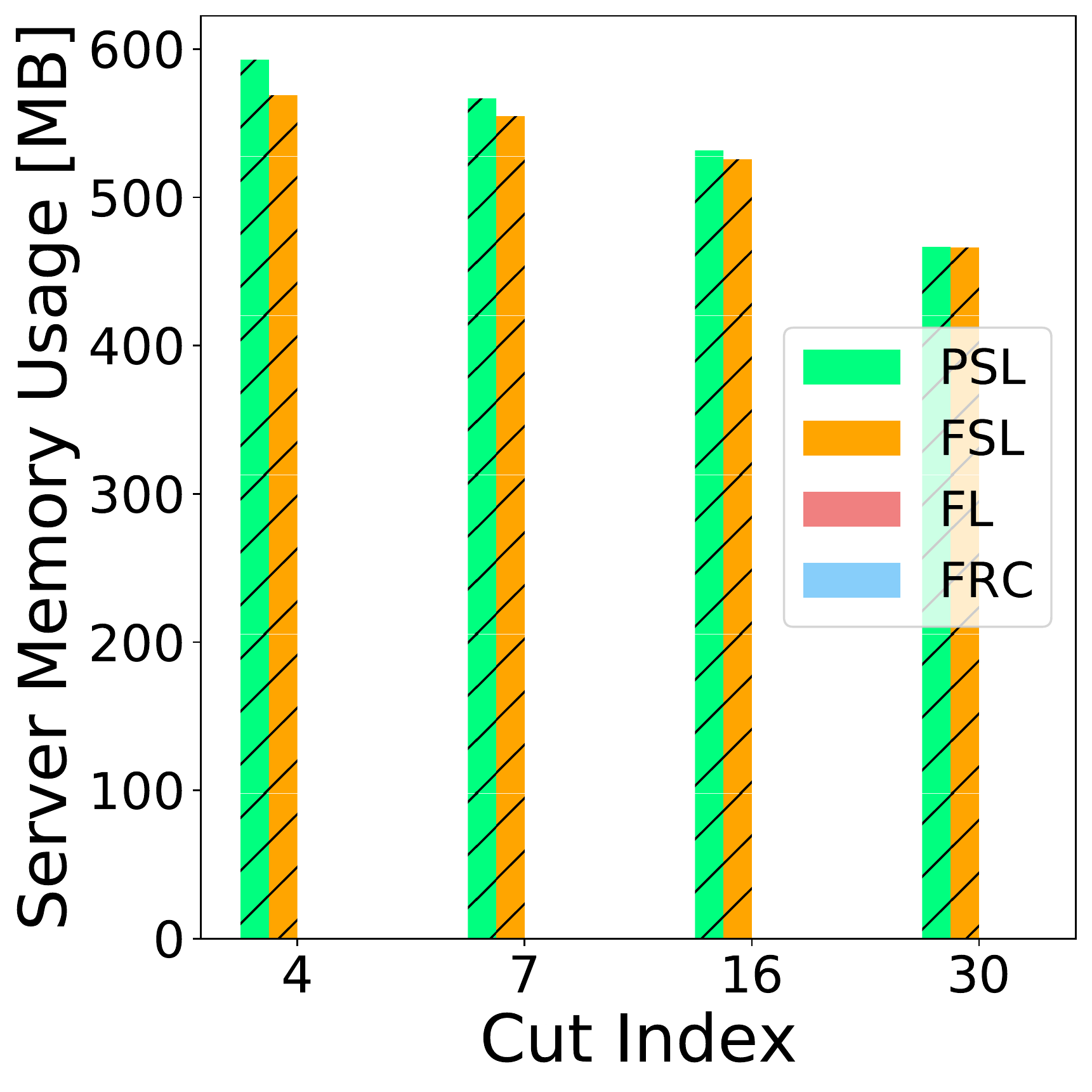}
    \caption{}
    \label{VGG16+Cifar-10-Memory-b}
    \end{subfigure}
    \caption{FSL's server has lower memory demand compared to PSL, and FSL's client has lower memory demand compared to FRC/FL.
    \textit{VGG + CiFar10}: (a) Client and (b) Server memory demand (sum of model weights size and outputs at each layer). Batch size is 32 and each image was resized to (3,32,32). 
    }{}
    \label{VGG16+Cifar-10-Memory}
\end{figure}

The two plots in Figure~\ref{VGG16+Cifar-10-Memory} show the memory demands computed at the client and the server for each architecture.
The x-axis shows the Cut Index and the y-axis represents the corresponding expected memory usages in MB.
Note that FL and FRC do not split the NN, so their memory demands are only shown in the left plot.

As shown in Figure~\ref{VGG16+Cifar-10-Memory-a}, the sizes of each client NN's weight and \zztbd{the results at each layer are the same in FSL and PSL.}
Since FRC and FL compute the full NN in the client during training, they require more than five times the memory, for Cut Index 4 to 30.
Figure~\ref{VGG16+Cifar-10-Memory-b} instead shows that the server memory demand decreases with the cut index, as expected.
However, notice that PSL server need more memory to hold intermediate data from different clients.

\noindent
{\bf
Memory  usage of FSL compared to other systems.} To conclude, we found that FSL's servers are lightweight compared to the PSL system. Consequently, state management would be easier in FSL. Also, FSL's clients are lightweight compared to FRC and FL, during training, so they are more suitable at edge.

\vspace{-2mm}
\subsubsection{Learner Accuracy Evaluation}\label{accu-eval}
\begin{figure*}[t]
\captionsetup[subfigure]{justification=centering}
    \centering
    \begin{subfigure}[t]{0.235\textwidth}
    \includegraphics[width=\textwidth]{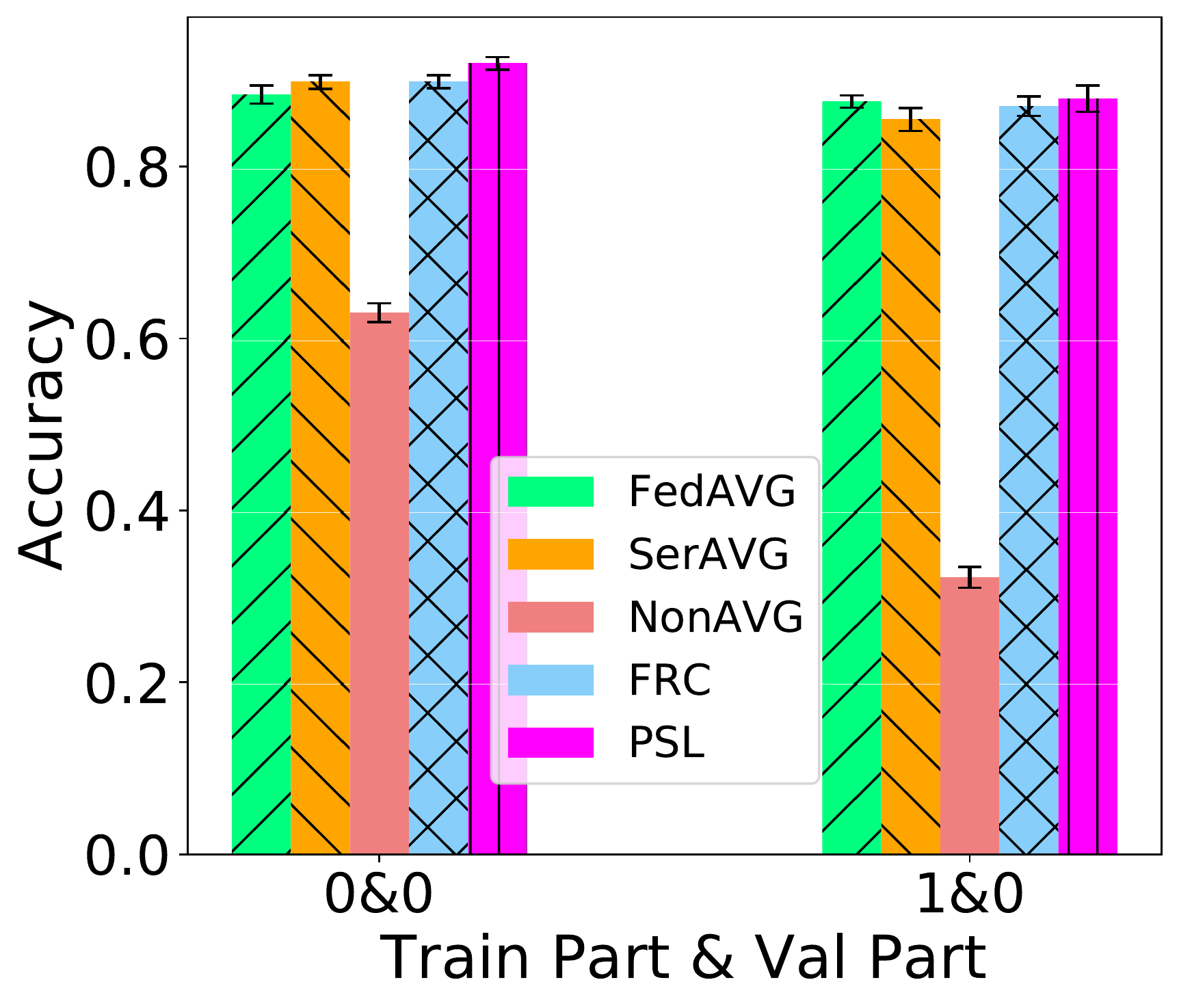}
    \caption{Accuracy (Cut Index = 3)}\label{lenet+mnist_avg_in_server_eval_3}
    \end{subfigure}
    \begin{subfigure}[t]{0.235\textwidth}
    \includegraphics[width=\textwidth]{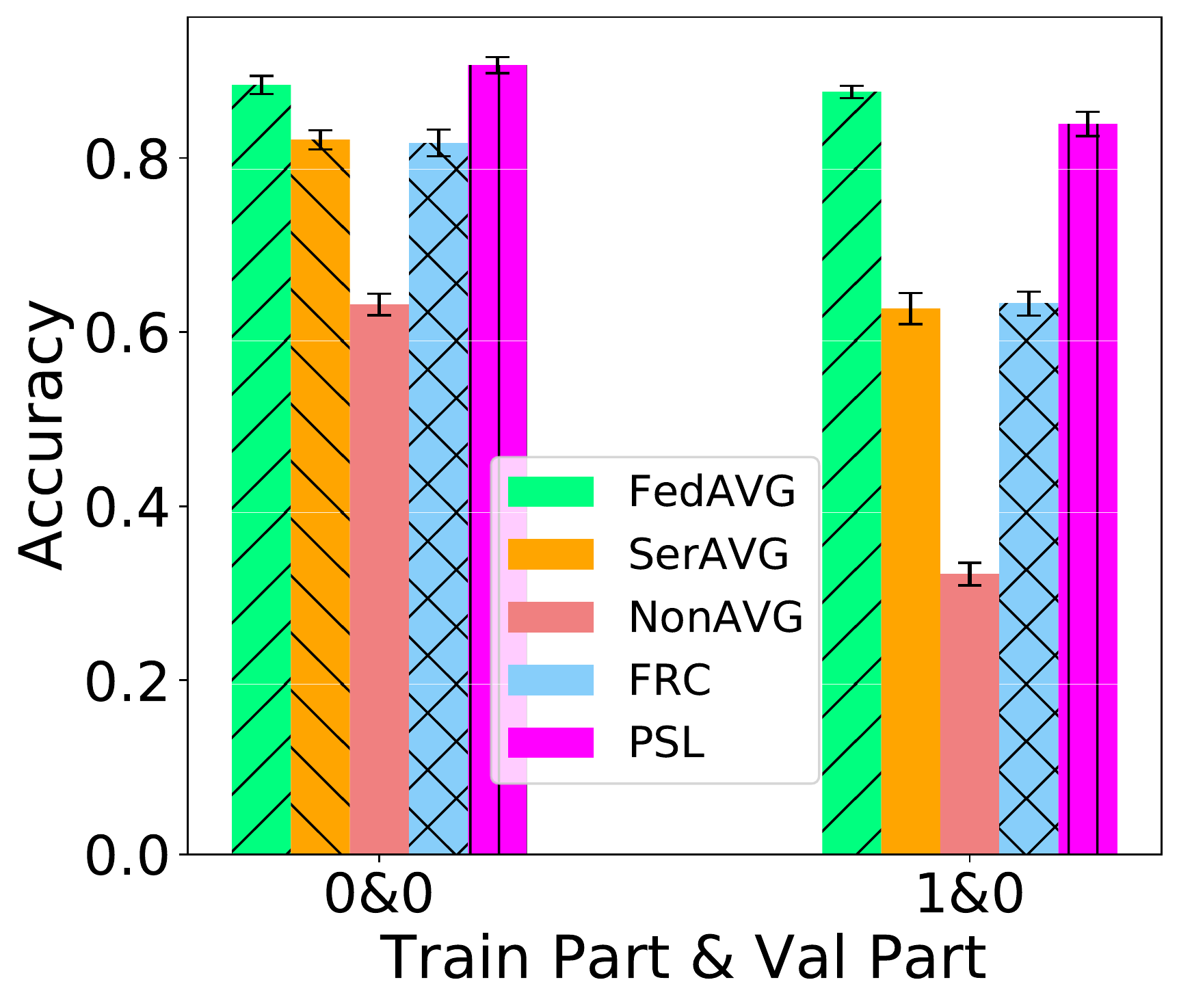}
    \caption{Accuracy (Cut Index = 5)}\label{lenet+mnist_avg_in_server_eval_5}
    \end{subfigure}
    \begin{subfigure}[t]{0.235\textwidth}
    \includegraphics[width=\textwidth]{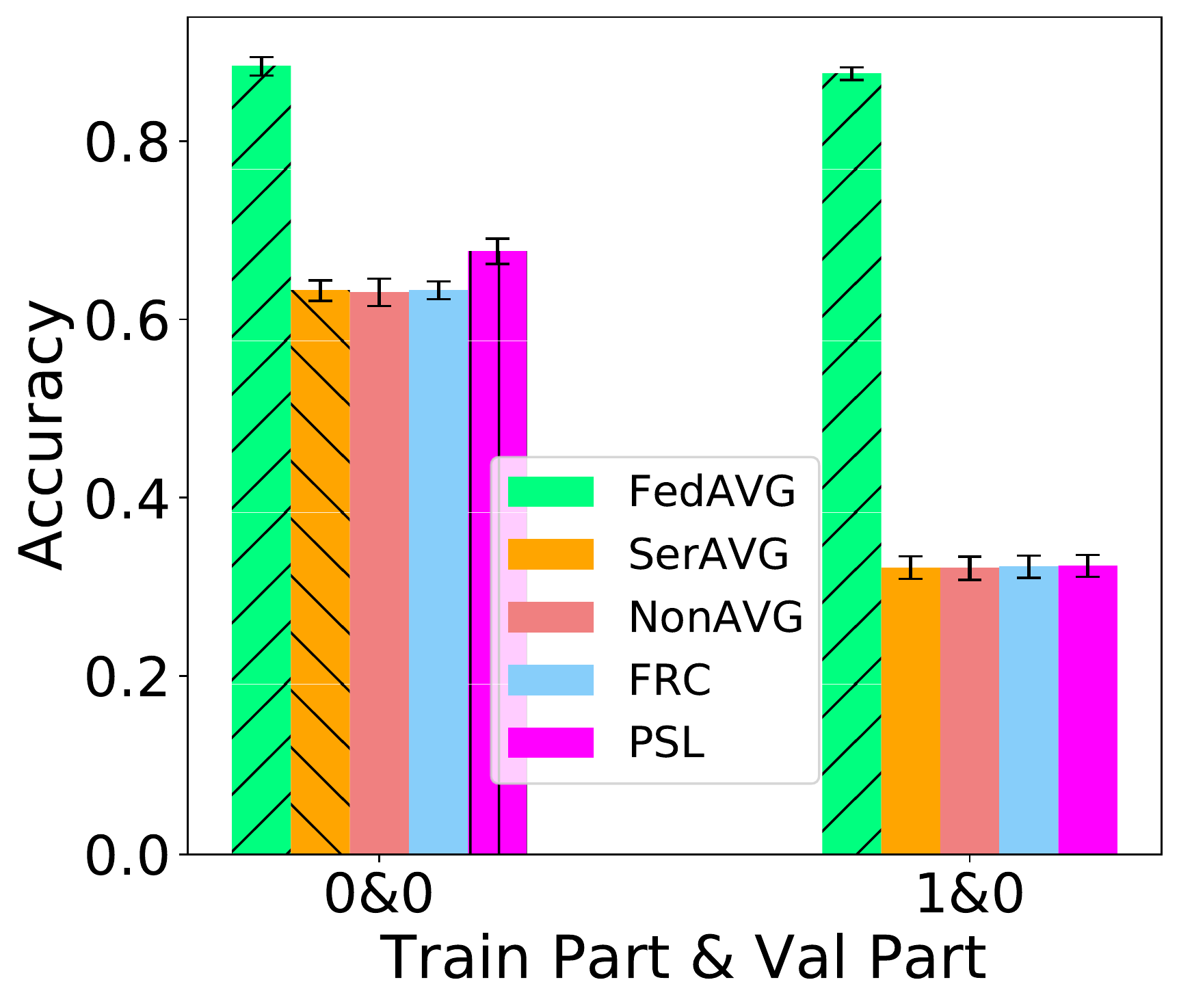}
    \caption{Accuracy (Cut Index = 7)}\label{lenet+mnist_avg_in_server_eval_7}
    \end{subfigure}
\hspace{2mm}
\begin{subfigure}[t]{0.23\textwidth}
    \includegraphics[width=\textwidth]{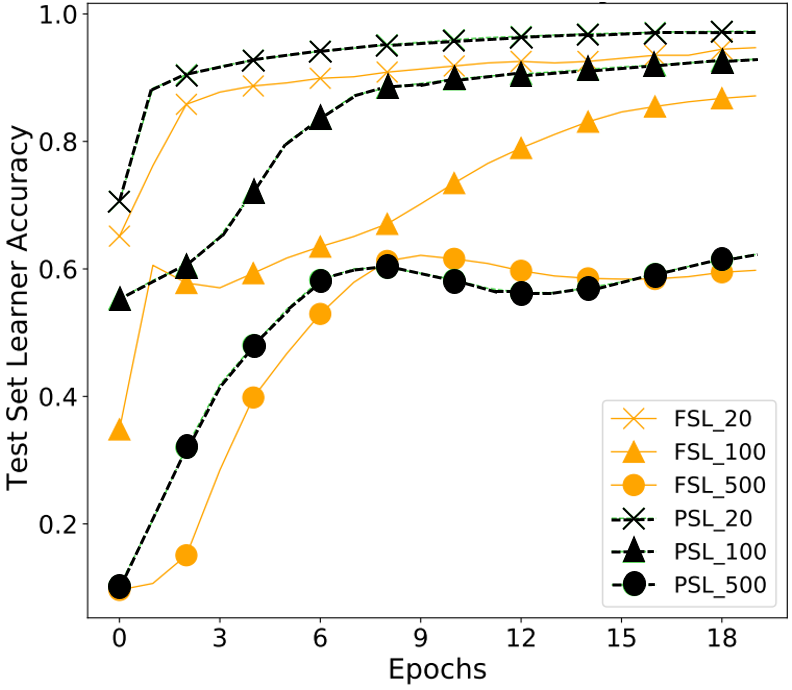}
    \caption{Accuracy (different number of clients)
    }{}
    \label{diff_num_client}
\end{subfigure}
\caption{
In plots~(a),~(b)~and~(c), the accuracy of SerAVG is better than NonAVG but lower while mostly close to FedAVG. 
SerAVG: Average the server NN's weights; NonAVG: Each pair trains on its own; FedAVG: Average the complete NN's weights.
In plot~(d), the FSL and PSL accuracy is similar. 
}

\end{figure*}

In this subsection, we focus on evaluating the convergence of SerAVG, proposed and detailed in Section~\ref{POFSL}.
Unlike FedAVG in Federated Learning, SerAVG averages the server NN weights. 
We begin by discussing the correctness of SerAVG, comparing the accuracy of SerAVG PSL, and FedAVG (FL). 
Then, we  compare the convergence rates based on different source data sizes and Cut Indexes.

\noindent
\textbf{SerAVG Evaluation}:\label{avgraging-eval}
In this experiment set, we evaluate if SerAVG can enhance the accuracy of every model joining the training process, given that the source data at clients are non-IID distributed. Our results are reported in Figures~\ref{lenet+mnist_avg_in_server_eval_3},~\ref{lenet+mnist_avg_in_server_eval_5}~and~\ref{lenet+mnist_avg_in_server_eval_7}.
To train the MNIST model, we split the training set in two parts, part 0 and part 1.
The two parts represent two \zztbd{collections of skewed data sources at client geolocations.}
We then let each part include data corresponding to half of the labels in the MNIST dataset.
We then split the MNIST test set into two parts in the same way.
Note that the aforementioned way of splitting training set and test set is an extreme \zztbd{Non-IID case. }
For example, a model trained on part 0 of the training set has no knowledge of the labels in part 1 of training set. 
To assess how well the systems may learn and predict on Non-IID data on more realistic data distributions, we further add $10\%$ samples uniformly and randomly, selected from the complete MNIST dataset to all the four parts \zztbd{of datasets} so that the model trained on either part of the training set may be able to classify labels in the other partition.
Consider Figures~\ref{lenet+mnist_avg_in_server_eval_3},~\ref{lenet+mnist_avg_in_server_eval_5}~and~\ref{lenet+mnist_avg_in_server_eval_7}.
The x-axis represents the data partitions used in training and validation, \zz{i.e.,\ $0~\&~1$ means data part 0 was used in training and part 1 was used during the validation phase.}
The y-axis shows the validation set accuracy.
From left to right, the accuracy decreases when the server/shared NN for FSL, PSL, and FRC is shallower.
When the Cut Index is 3, SerAVG, FedAvG, and PSL perform equally well. 
When the Cut Index is greater than 3, SerAVG is worse than FedAVG and lower but close to PSL.
We conclude that SerAVG can enhance the accuracy in the Non-IID source data setting, while worse than FL and PSL.

\zz{
Note that FedAVG averages all weights, while FSL never shares the client’s weight in the SerAVG setting. 
Thus, FSL clients cannot benefit from the gradients calculated at other clients and that may lead to lower accuracy.
Comparing PSL and FSL, 
the PSL server optimizes for minimal loss using all clients’ batch output.
On the other hand, SerAVG averages the trained weights on each server heuristically based on FedAVG. Thus, SerAVG will have lower accuracy compared with PSL.
And each client NN trained with a Non-IID dataset can extract little features from the other Non-IID dataset.
}

\begin{figure*}[t]
\captionsetup[subfigure]{justification=centering}
    \centering
    \begin{subfigure}[t]{0.245\textwidth}
    \includegraphics[width=\textwidth]{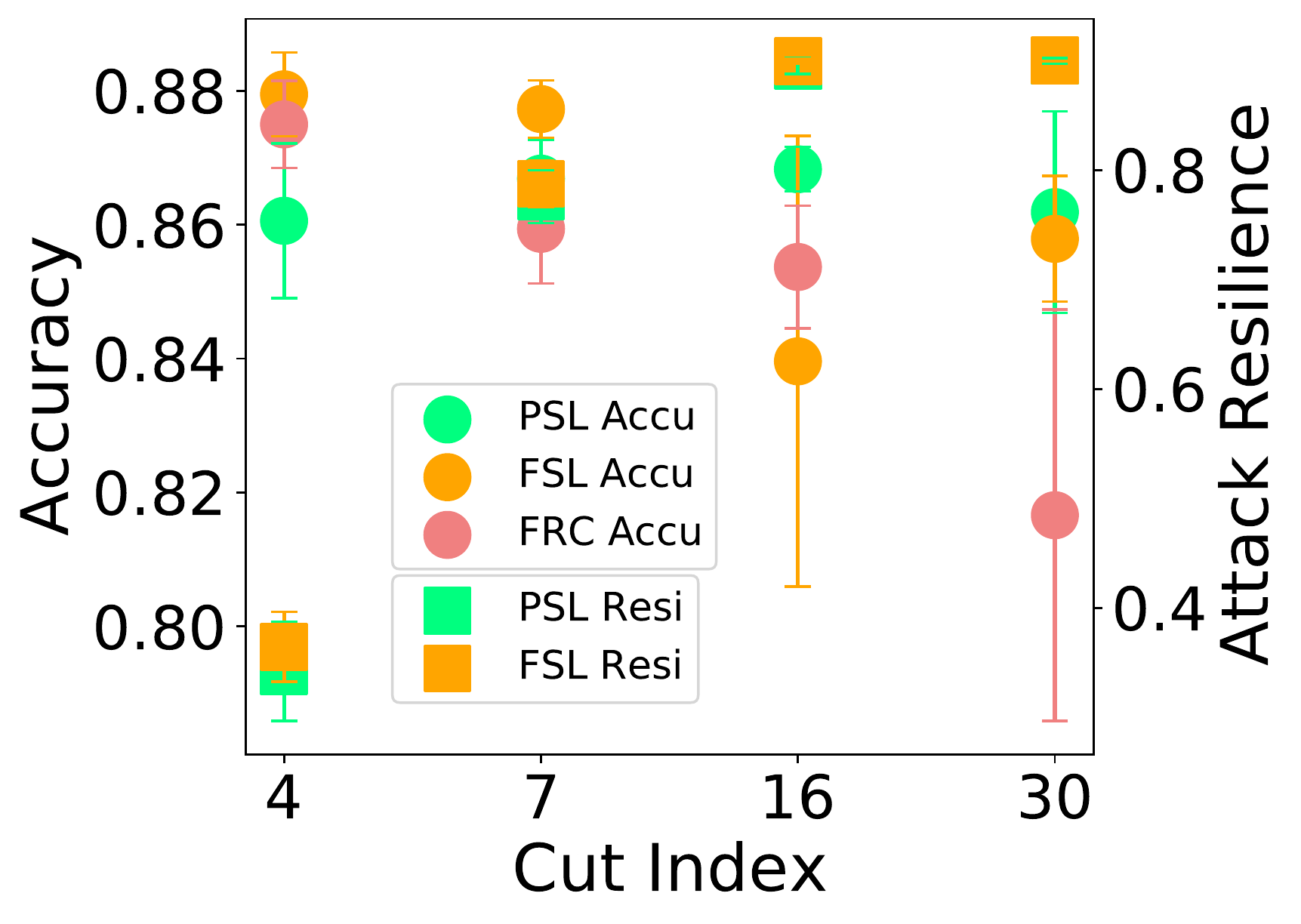}
    \caption{Privacy Oblivious \\ (VGG16+Cifar-10)}
    \label{accu&resi_CIFAR10_VGG16_oblivious}
    \end{subfigure}
    \begin{subfigure}[t]{0.24\textwidth}
    \includegraphics[width=\textwidth]{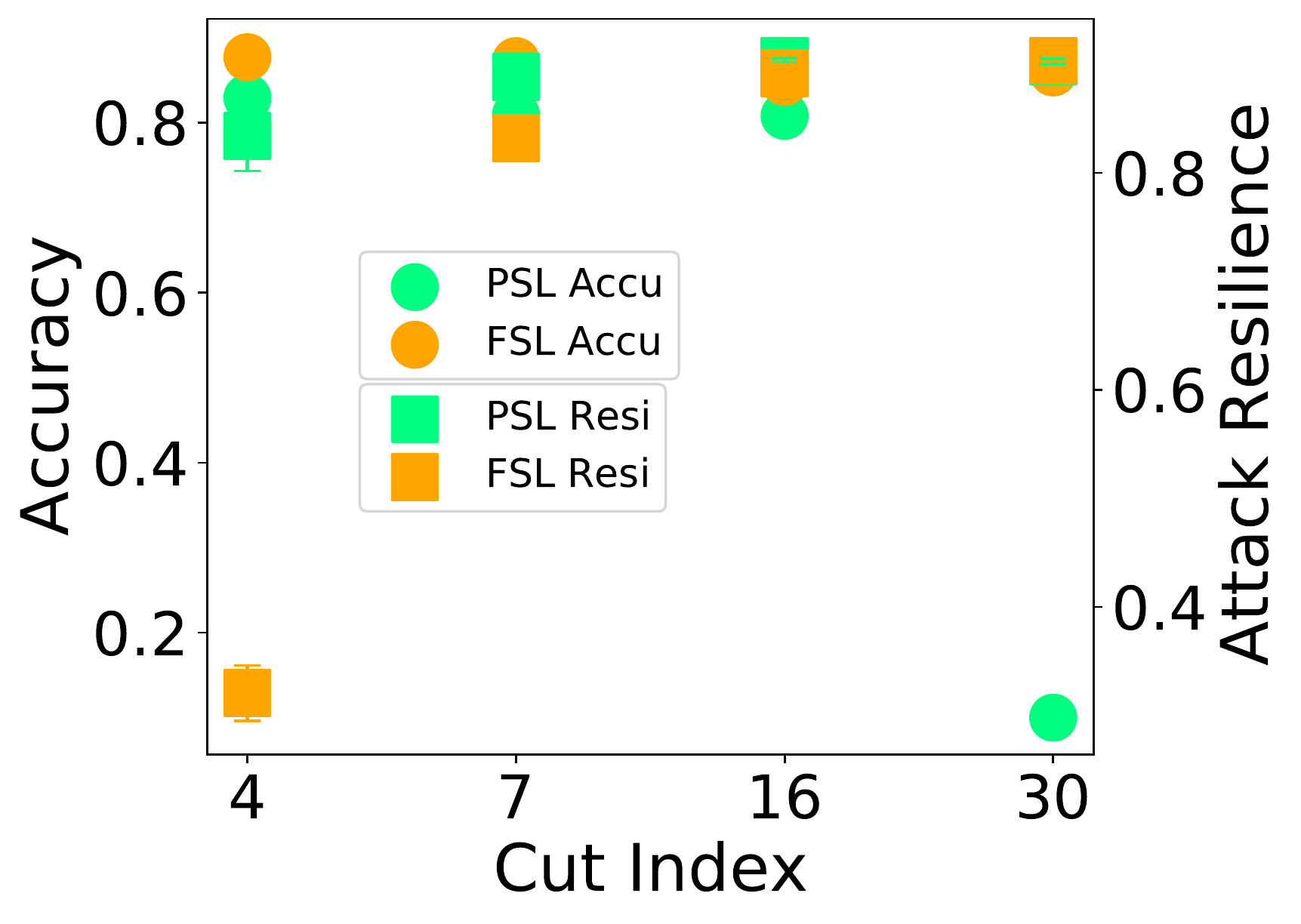}
    \caption{DC 1 \& Muiltiplier=2.0 \\ (VGG16+Cifar-10)}
    \label{accu&resi_CIFAR10_VGG16_aware_20}
    \end{subfigure}
    \begin{subfigure}[t]{0.245\textwidth}
    \includegraphics[width=\textwidth]{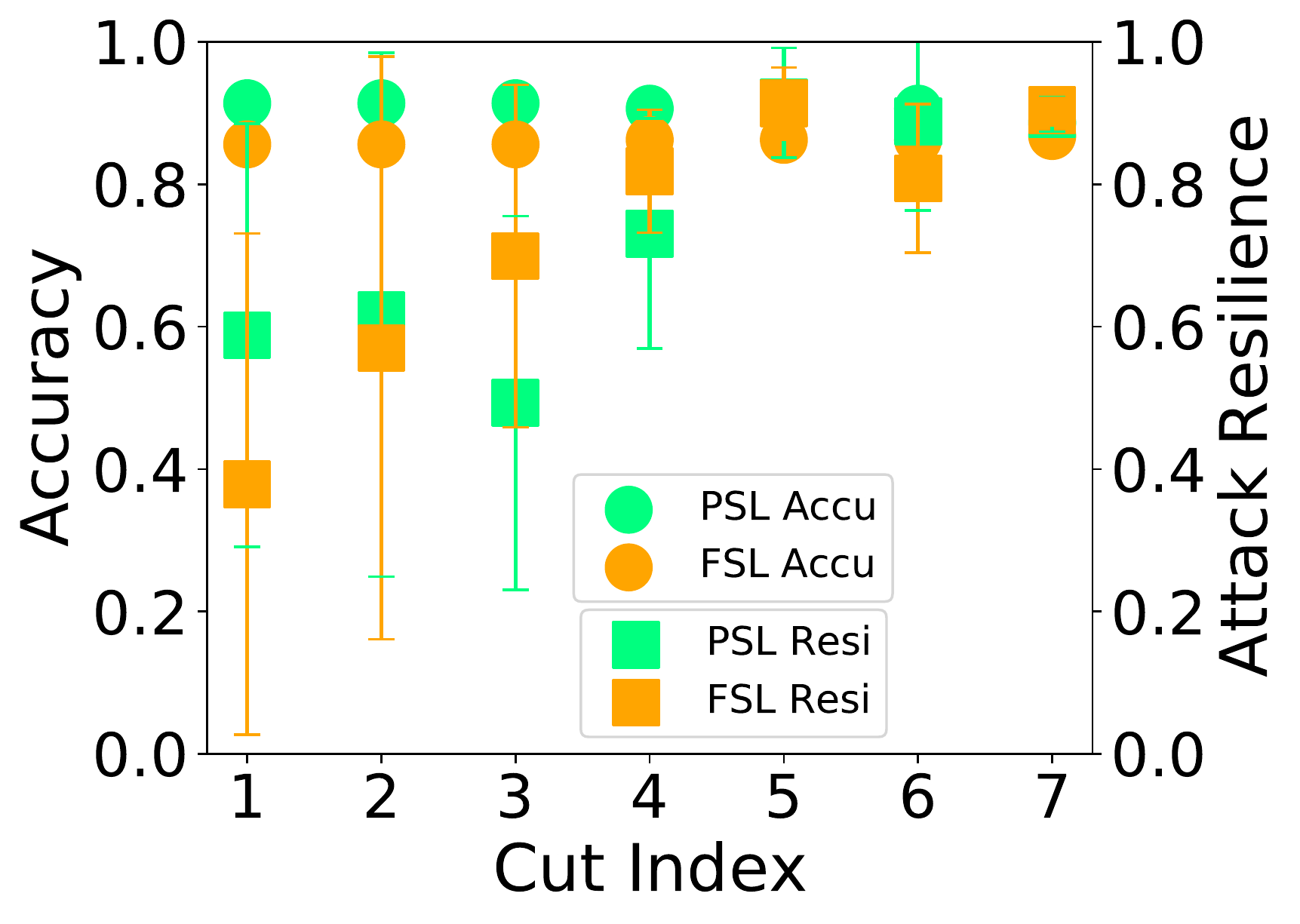}
    \caption{Privacy Oblivious \\ (Lenet+MNIST)}
    \label{accu&resi_MNIST_LENET_aware_nop}
    \end{subfigure}
    \begin{subfigure}[t]{0.245\textwidth}
    \includegraphics[width=\textwidth]{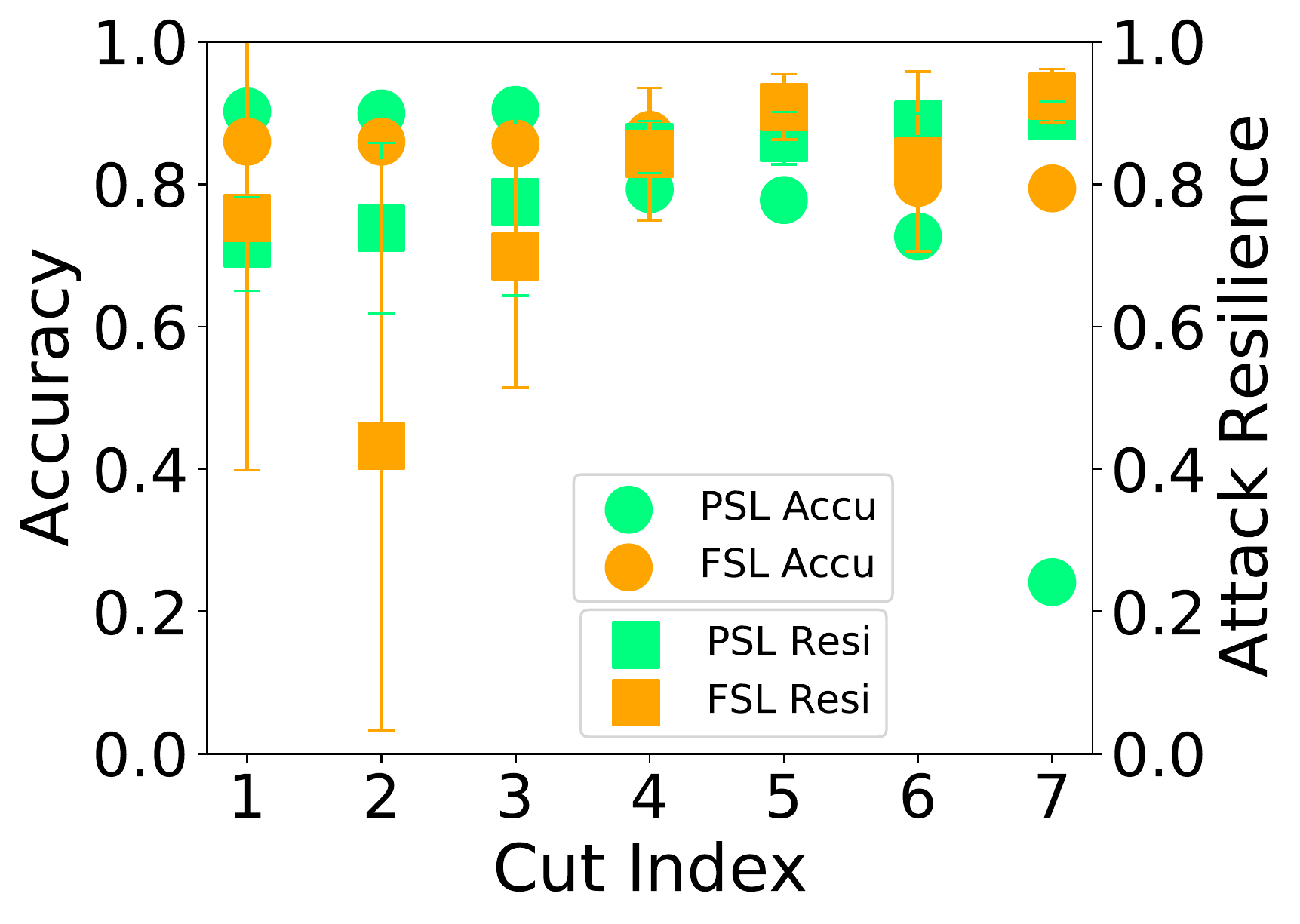}
    \caption{DC 1 \& Multiplier=0.3 \\ (Lenet+MNIST)}
    \label{accu&resi_MNIST_LENET_aware_03}
    \end{subfigure}
    \caption{Accuracy and Attack Resilience for Privacy Oblivious and Privacy Aware Architectures based on Cut Index. Note: DC 1 in captions refer to $DC\_Frequency=1$. 
    These figures show that to reach high accuracy and attack resilience, the Cut Index cannot be too big or too small and loss multiplier is another way to enhance attack resilience. Furthermore, loss multiplier can be more practical than DC\_Frequency, since it doesn't introduce overhead to the training steps.
    }{}
    \label{multipliers}
    \vspace{-2mm}
\end{figure*}

Note also a similar but smoother drop in accuracy based on Cut Indexes with Independent and Identically Distributed (IID) partitioned CIFAR10 dataset and VGG16 NN in Figure~\ref{accu&resi_CIFAR10_VGG16_oblivious}.
The two experiments with MNIST and CIFAR10 datasets suggest that SerAVG can preserve similar accuracy patterns even when applied to different NN models and Cut Indexes.

\noindent
\textbf{Tradeoff between resource demand and accuracy.}
In this experiment we evaluated the accuracy of FSL with small resource demand, i.e., when the size of the input data is limited, and the client model runs on limited resources.
We found that with LENET and MNIST, the validation dataset  can reach an accuracy range of $87\%$ to $93\%$, as long as each client has enough data to train the machine learning model.
By varying the number of clients, we quantified the expected drop in accuracy for both architectures.
Our results are shown in  Figure~\ref{diff_num_client}.
The x-axis indicates the index of epochs, and the y-axis shows the corresponding test set accuracy after a certain number of epochs.
\zztbd{The dataset is IID among 20, 100, and 500 clients to study how the data size affects convergence.}
In Figures~\ref{lenet+mnist_avg_in_server_eval_3}, \ref{lenet+mnist_avg_in_server_eval_5}, \ref{lenet+mnist_avg_in_server_eval_7} and \ref{accu&resi_CIFAR10_VGG16_oblivious}, we also noted that FSL (implementing the SerAVG mechanism) keeps high accuracy when the Cut Index is small, allowing deployments over resource limited device given high performance requirement.
\noindent
We expect a higher accuracy for both architectures with more effort in tuning the hyper-parameters,  given prior results in similar contexts~\cite{RandomSearch, hyperband}. 
While our accuracy results show $95\%$ confidence intervals, parameter tuning 
is out of the scope of this paper.

\vspace{-2mm}
\subsection{Evaluation Results for Privacy-Aware FSL}\label{expr-PAA}
In this section we present the evaluation results of our privacy-aware FSL (PAFSL) architecture and show that it can provide \zztbd{certain} privacy guarantees.
FSL clients do not share the source data and model weights, so adversaries cannot directly access the source data or reconstruct them with the model weights using model inversion attacks~\cite{model-inversion-attack-1, model-inversion-attack-2}.
However, when compared with Federated Reconstruction (FRC)~\cite{federatedreconstruction} which trains a complete model at the client, FSL still sends the intermediate data through a network to complete the forward and backward propagation between clients and (edge) servers. An adversary could use such data to reproduce the source data, e.g., through an Autoencoder~\cite{autoencoder} NN, 
trained with certain dataset, in Section~\ref{sec:attack}.

To assess how our approach mitigates such vulnerabilities, we first introduce the evaluation setup, the design and usage of the attacker Autoencoder NN, and our experiment methodology.
Then, we discuss the results of different \zztbd{privacy} approaches, i.e., NoPeek~\cite{nopeekarticle} \zztbd{and} the Client Based Privacy Approach (CPA), \zztbd{and privacy level of different ways of paritioning the NN.}
NoPeek solves a multi-objective optimization problem of two loss functions, i.e., one maximizes accuracy, and the other maximizes the differences between source images and intermediate data. 
For CPA, we evaluate CPA-DC and CPA-DP.
CPA-DC (Section~\ref{CPA}) \zztbd{optimize the two loss function in NoPeek alternatively.}
CPA-DP applies a DP-SGD\cite{DLwithDP} algorithm in the clients.
Finally
we evaluate the privacy guarantee of different partitioning of client NN and server NN \zztbd{motivated by Section~\ref{NPS}.}
%
We conclude this section by presenting results that demonstrate the high resilience to privacy attacks  of our proposed FSL and the advantages in training efficiency.

\vspace{-2mm}
\subsubsection{Evaluation Settings}\label{sec:PA-setup}

\zztbd{
Our Privacy-Aware FSL and PSL extend our Privacy-Oblivious version by adding the CPA. 
Partitioning the NN was made easy by considering only sequential NNs (e.g., LeNET and VGG16). 
We tested both systems with the same image classification workloads (e.g., MNIST and CIFAR10) on the same hardware (i.e., NVIDIA RTX6000 and NVIDIA V100, respectively) as the Privacy-Oblivious setting.
}


%

\vspace{-2mm}
\subsubsection{Setup the Attacker's Auto-Encoder Neural Network}\label{sec:PA-autoencoder}

In this subsection we explain how we define our privacy attack, given the assumptions mentioned in Section~\ref{sec:attack}. To understand how the privacy attack works, it is useful to recall how the Autoencoder NN that we used work. In particular, the Autoencoder is composed of two parts: an encoder and a decoder. 
We need a dataset similar to the source images to train the Autoencoder NN.

While the encoder uses convolutional layers to extract latent variables from its input dataset, the decoder uses the last layer's activation function outputs from the encoder and transposes the convolutional layers to reproduce the input dataset of the encoder.
Consequently, the encoder NN structure is the same as the client NN structure, and the attacker is the decoder NN.
The $i^{th}$ layer of encoder would be the $i^{th}$ last layer of decoder, with transposed convolutional layer. 
We assume the attacker's decoder structure strictly mirrors the client NN structure.
Thus, we expect better attack resilience for SL-based systems in the real world.


\vspace{-2mm}
\subsubsection{Privacy Evaluation: Methodology}\label{privacy-eval}\label{sec:PA-methodology}
In this subset of our evaluation, we want to show \zztbd{both CPA and carefully designed ways of partitioning NN provide high attack resilience while preserving high accuracy, based on different datasets, NNs, for split learning based systems.}

Each experiment includes trials initializing the Autoencoder's weights and data-loaders with different random seeds.
In each trial,
the attacker used a dataset containing similar features to the learner's source dataset.
To reconstruct MNIST, we selected EMNIST~\cite{EMNIST}, and for CIFAR10 we selected the CIFAR100~\cite{cifar10}.
The EMNIST dataset contains hand-written characters instead of numbers in MNIST, so features like lines and curves are the same and attacker can decode those activation function outputs. 
Similarly, the CIFAR100 dataset contains 100 classes of RGB images instead of the 10 classes in CIFAR10, so the common features, e.g. classifying cat or dog, can be used to reconstruct with
the activation function outputs from the CIFAR10 dataset.

For each trial of the experiment, we first let the attacker learn to reproduce her datasets.
Based on the MSE loss, {\em i.e.}\ a loss function that measures how different the original and reproduced images are, the attacker updates her weights in each epoch.
After 20 epochs, we used the decoder to reproduce the learner's dataset from the intermediate data.

When an autoencoder NN is trained, we train a new classifier with the same NN structure and source data as the learner to classify the reproduced images for 20 epochs.
The mean and standard deviation of this classifier's {\em attack resilience} $\tau~$ in section \ref{resilience_tau}.
We show the results comparing the four systems, i.e., POFSL, POPSL, PAFSL, and PAPSL, with different CPAs and Cut Indexes.
Then, we evaluate the trade-off between accuracy and attack resilience for FSL and PSL.

\vspace{-1mm}
\subsubsection{Privacy Evaluation using NoPeek}\label{peval-NoPeek}\label{sec:PA-nopeek}

As we illustrated in Section~\ref{RW:SDP}, NoPeek solves a multi-objective optimization problem that takes in the source data and intermediate data to maximize the difference with a Distance Correlation (DC) loss function, as well as the prediction and labels to maximize the accuracy.
To solve such optimization problem, NoPeek has to share the value of the loss over an network which may cause vulnerability or added complexity of maintaining the gradient graph.
As shown in Table~\ref{table_attack_resilience}, this approach has both high attack resilience (i.e., 97\% for PSL and 98\% for FSL) and high learner's accuracy (i.e., 97\% for PSL and 96\% for FSL) when trained for the same number of epochs and clients as the Privacy Oblivious experiment with the MNIST dataset and LENET NN. 

\begin{table}[!h]
\centering
\begin{tabular}{c | c | c}
 cases & PSL & FSL
\\ \hline  
 attack\_resilience($\tau$) & 0.9733 & 0.9837
\\ \hline  
 learner\_accuracy & 0.9702 & 0.9614
\\ \hline  
\end{tabular}
\caption{\zztbd{NoPeek stats with 20 clients training $20$ epochs.}}
\label{table_attack_resilience}
\end{table}

\vspace{-6mm}
\subsubsection{Evaluation Result using Client-Based Privacy Approach via Distance Correlation}\label{sec:PA-CPA-DC}
To mitigate the drawbacks of the loss value sharing, we consider a new approach that prevents transmitting data outside clients, improving upon NoPeek.
We optimize for the similar two objectives in NoPeek alternatively in the Client-Based Privacy Approach (CPA) via DC.
As shown in 
Equation~\ref{CPADC_loss_function}, there are \textit{DC Frequency~(F)} and \textit{Loss Multiplier~(m)} to evaluate. 
\textit{DC Frequency~(F)} defines how many times the DC loss function is optimized after the loss function for accuracy is optimized once.
\textit{Loss Multiplier~(m)} is applied to the loss function result but has the equivalent effect of multiplying the learning rate by a factor $m$.
These two parameters control how different the intermediate data and source data will be,
by changing the frequency of optimizing the DC loss and by changing the learning rate of gradients applied during that optimization, respectively. 

\begin{figure}
    \centering
    \begin{subfigure}[t]{0.48\textwidth}
    \includegraphics[width=\textwidth]{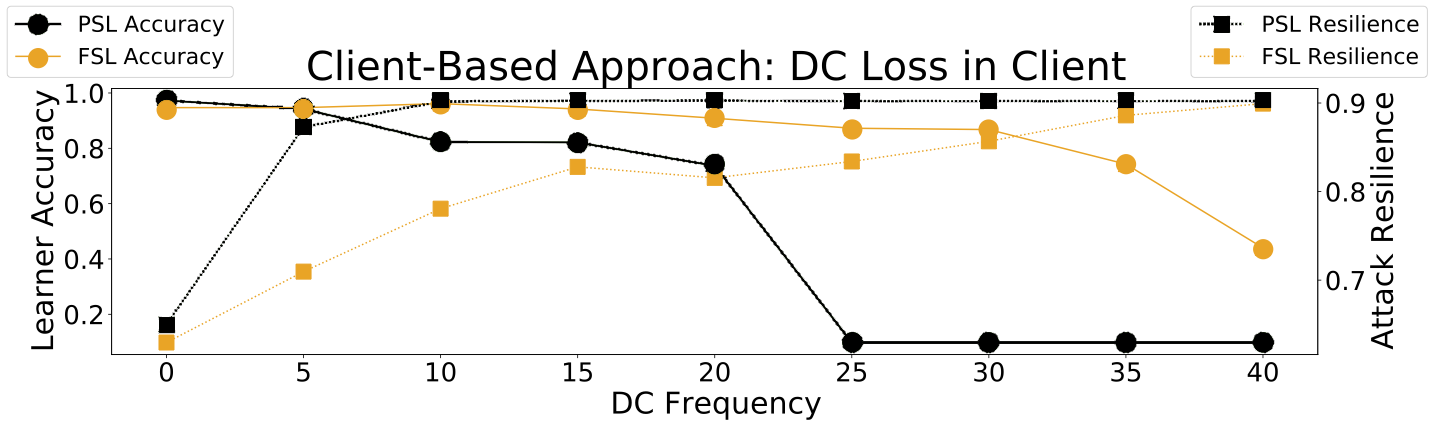}
    \caption{CPA-DC (Multiplier=0.1)}
    \label{CPA-DC-Results}
    \end{subfigure}
    \begin{subfigure}[t]{0.48\textwidth}
    \includegraphics[width=\textwidth]{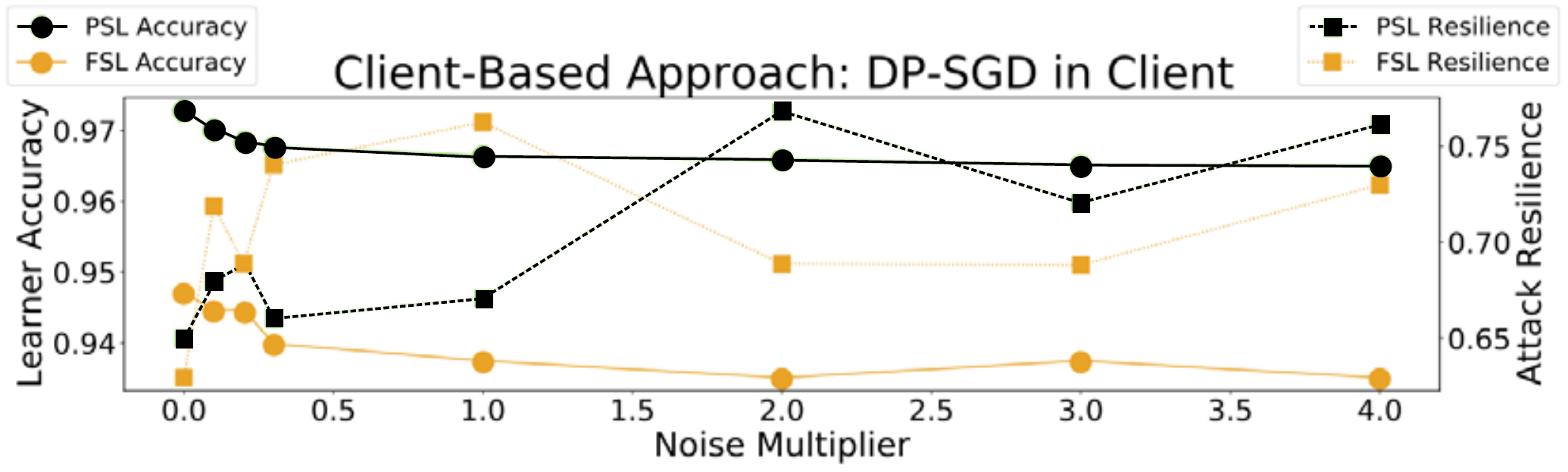}
    \caption{CPA-DP}
    \label{CPA-DP-Results}
    \end{subfigure}
    \begin{subfigure}[t]{0.48\textwidth}
    \includegraphics[width=\textwidth]{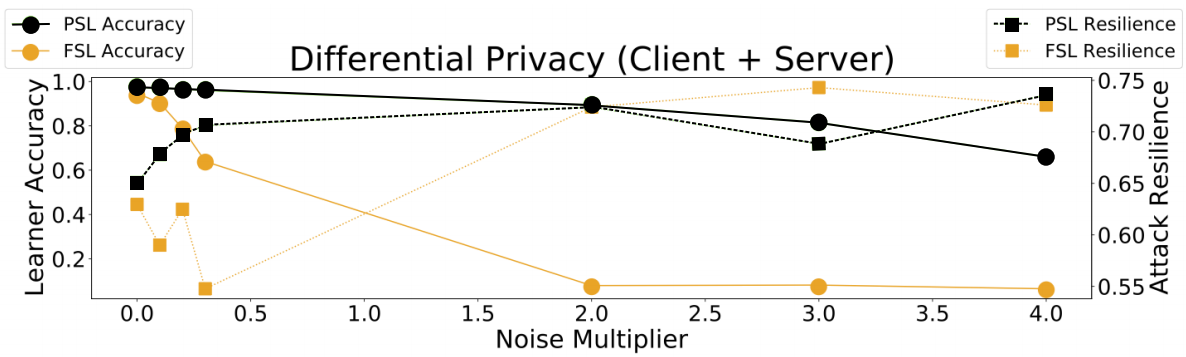}
    \caption{DP-SGD in Complete NN}
    \label{DP-SGD-Results}
    \end{subfigure}
    \caption{LeNet+MNIST: Learner accuracy and 
    attack resilience ($\tau$) with 20 clients and Cut Index of 3 for Client-Based Privacy Approaches (via DC (a) and via DP (b)) and DP-SGD on the global learner model (c). Our FSL with both client-based policies guarantee high-level of privacy and accuracy.}{}
    \label{20th_epoch_accu_plot}
\end{figure}

\zz{We note multiple tradeoffs in CPA-DC. First,} CPA-DC is not as training time-efficient as NoPeek. NoPeek can optimize its two objectives simultaneously while CPA-DC has to solve them sequentially.
However, we consider that NoPeek transfers more information than necessary  over a network. 
\zz{
Second, there are tradeoffs for \textit{DC Frequency~(F)} and \textit{Loss Multiplier~(m)}.
Increasing \textit{F} adds more epochs to optimize for DC loss, so the attack resilience would be higher at the expense of a longer training time.
Moreover, we introduced \textit{m}, so that we can keep a small \textit{F} and only increase \textit{m}, which reduces training time while maintaining attack resilience.
We multiply larger \textit{m} by the DC loss, similar to increasing the gradient descent step size. 
Thus, we need less DC epochs to maintain the attack resilience, given a larger \textit{m}, while the DC loss becomes less accurate.
Based on the discussion, intuitively we expect that a large \textit{m} combined with \textit{F} of $1$ can balance between training time efficiency and DC loss gradients' accuracy.
These two parameters should be carefully designed in a production environment.
}
We first experimented MNIST classification with different DC Frequencies with a constant loss multiplier of 0.1 (equivalent to reducing the learning rate by $10\%$) and studied the tradeoff between accuracy and attack resilience, as shown in Fig.~\ref{CPA-DC-Results}.
The x-axis represents the DC Frequency, the left y-axis shows the learner accuracy and the right y-axis shows the corresponding attack resilience.

From the top plot of Figure~\ref{20th_epoch_accu_plot}\footnote{A DC Frequency of zero corresponds to POFSL and POPSL without the privacy-aware approaches.}, as DC Frequency is increasing, for both Privacy Aware FSL (PAFSL) and Privacy Aware PSL (PAPSL) systems, the attack resiliency increases and the learner accuracy decreases, as expected.

Notice that PAFSL achieves better accuracy and good resilience for most DC Frequency values.
For DC Frequency from 10 to 20, given that the attack resilience of PAFSL and PAPSL are close within $10\%$ difference, PAFSL achieves more than $90\%$ accuracy.
From 25 to 35, PAPSL does not learn any features while PAFSL still has about $80\%$ accuracy.
When DC Frequency is five, the PAPSL has an advantage over PAFSL, with close accuracy, and PAPSL has around $20\%$ more attack resilience.\\
{\bf The result shows that PAFSL with CPA-DC is easier to tune for high accuracy and attack resilience.} {\it Within wider domain of DC Frequency, PAFSL has higher accuracy and good attack resilience compared to PAPSL.}
This is because of the learning rate (step size) in SerAVG and PSL.
The server weight update rule of PSL is shown in Equation~\ref{PSL_server_update}.
\vspace{-1 mm}
\begin{equation}\label{PSL_server_update}
    W_g^{t+1} = W_g^{t} - \eta\sum_{i=1}^{N_{C}} \dfrac{\partial g(f(x_i))}{\partial W_g} 
\end{equation}
The server weight update rule of FSL is shown in Equation~\ref{FSL_server_update}.

\vspace{-5mm}
\begin{equation}\label{FSL_server_update}
\begin{split}
    W_g^{t+1} &= \frac{\sum_{i=1}^{N_{C}}(W_g^{t} - \eta \dfrac{\partial g(f(x_i))}{\partial W_g})}{N_{C}}\\
    &= W_g^{t} - \frac{\eta}{N_{C}}\sum_{i=1}^{N_{C}} \dfrac{\partial g(f(x_i))}{\partial W_g}, 
\end{split}
\end{equation}
where $W_g^{t}$ indicates the weights in the server at iteration $t$, $g$ is the server NN, $f$ is the client NN, $x_i$ represent the i-th batch of data, $N_{C}$ is the number of clients, and $\eta$ is the step size.
Intuitively, since PSL has a larger step size, its server NN can be confused quicker than FSL servers by the intermediate data.
Moreover, the confused server NN can further confuse the client NN. 
It justifies our observation that PSL's accuracy and attack resilience become unstable quickly when increasing the \textit{DC Frequency~(F)}. Therefore, we conclude that {\it FSL is easier to tune compared to PSL.} 


\zz{In Figures~\ref{accu&resi_CIFAR10_VGG16_aware_20} and~\ref{accu&resi_MNIST_LENET_aware_03}, 
with a fixed \textit{DC Frequency~(F)}, we show the accuracy (left y-axis) and attack resilience (right y-axis) based on different \zz{\textit{Loss Multiplier~(m)}} \zz{for different models and datasets}.}
Furthermore, we compared the privacy oblivious cases (Figure~\ref{accu&resi_CIFAR10_VGG16_oblivious} and Figure~\ref{accu&resi_MNIST_LENET_aware_nop}), and the privacy aware cases at different Cut Indexes (x-axis).
\zz{As expected, increasing the \textit{Loss Multiplier~(m)} enhances attack resilience but reduces accuracy, especially when the client NN is deep.}

\zz{Overall our evaluation of CPA-DC shows good attack resilience and accuracy with a combination of small \textit{DC Frequency~(F)} and big \textit{Loss Multiplier~(m)}.
And our FSL has better accuracy and similar attack resilience to PSL.
We hence conclude that our CPA-DC can defend against our attacker model.}

\vspace{-2mm}
\subsubsection{Privacy Evaluation with Differential Privacy Approach}\label{eval-CPA-DP}\label{sec:PA-CPA-DP}
\zz{
The previous section has discussed the CPA-DC, but instead of DC there are other lightweight methods that can 
enhance the privacy guarantee which adds noise to the client NN while prevent depleting the client battery quickly.
In this section, we compare CPA-DP (using the popular DP-SGD~\cite{DLwithDP} algorithm inside clients) and CPA-DC.
Also, we show that na\"{\i}vely using DP-SGD in an FSL system would lead to low accuracy.
The implementation extends POFSL with a DP-SGD optimizer, provided by the Opacus~\cite{opacus} library.
This method would add normally distributed random noise to the gradients during backward propagation based on $noise\_multiplier$ $\epsilon$. 
This parameter controls the magnitude of the noise added.
Notice that DC generates the gradients in a specific direction to reduce correlation between intermediate data and source data in each \textit{Distance Correlation round}.
So we expect CPA-DP to have a worse level of privacy, given the same level of learner accuracy, compared to CPA-DC.
Thus, the focus of this section is to show that CPA can be applied with other privacy methods like DP, despite DP's worse privacy compared to DC.
}


We summarize the results of CPA-DP in Fig.~\ref{CPA-DP-Results}.
This plot shows the result when Cut Index equals 3.
The x-axis is the $noise\_multiplier$.
\vv{The attack resiliency of FSL and PSL with $noise\_multiplier > 0$ is consistently better by nearly $5\%$ than FSL and PSL with $noise\_multiplier = 0$.}
At the same time, the accuracy decreases by less than $1\%$ in either FSL or PSL from $noise\_multiplier = 0$ to $noise\_multiplier = 4$.

{\bf CPA is a general approach and can be customized with different methods to enhance attack resilience.}  The evaluation shows that CPA-DP can also improve attack resilience, while the learner accuracy does not change much.
Comparing with CPA-DC, both methods provide similar learner accuracy, but CPA-DC's attack resilience is higher.

We now compare CPA-DP against applying DP-SGD in both client NN and server NN.
Figure~\ref{DP-SGD-Results} shows high attack resilience, but the learner accuracy of FSL drops below $60\%$ when $noise\_multiplier \ge 0.3$.
Meanwhile, PSL shows a similar behavior as using CPA-DP.
So, CPA-DP is considered a better method for FSL to enhance its attack resilience than with DP-SGD applied in both clients and servers. 
The reason for FSL's lower accuracy under DP-SGD and high noise can be attributed
to its SerAVG.
After applying SerAVG, the distribution of the random noise in the server NN can be arbitrary, as shown in E.q., \ref{FSL_server_update_DP-SGD}, while the noise in the client NN stays intact.
Thus, the resulting complete NN in FSL may not converge.
\vspace{-3mm}
\begin{equation}\label{FSL_server_update_DP-SGD}
    W_g^{t+1} = W_g^{t} - \frac{\eta}{N_{C}}\sum_{i=1}^{N_{C}} (\dfrac{\partial g(f(x_i))}{\partial W_g} + e_{g_i}^{t})
\end{equation}

\vspace{-1mm}
The variable $e_{g_i}^{t}$ indicates the gaussian noise added for the $i$-th server NN at iteration $t$ according to $e_{g_i}^{t}\sim\mathcal{N}(0,\,(noise\_multiplier \times max\_gradient\_norm)^{2})\,$
\cite{opacus}.

\subsubsection{Evaluation Using Different ways of Partitioning NN}\label{eval-NPS}\label{sec:PA-NPS}
In Figure~\ref{accu&resi_CIFAR10_VGG16_oblivious}~and~Figure~\ref{accu&resi_MNIST_LENET_aware_nop}, the right y-axis shows the attack resilience, and the x-axis indicates the Cut Indexes.
Overall, if we have a deeper client NN (i.e., moving from smaller to bigger Cut Indexes), the attack resilience increases and accuracy decreases 
(consistent with the SerAVG Evaluation in Section~\ref{accu-eval}).
After adding more layers, the intermediate data would have less features from the source data, but only keeps those that can improve the classification \zztbd{accuracy.}
Thus, less features are preserved and the attacker's ability to reconstruct the source's data is hindered.

We also want to emphasize that with CIFAR10 workload, when there are 7 layers in the clients, the attack resilience reaches about $80\%$. 
We reach the same result with MNIST workload at the Cut Index of 4.
No extra privacy-aware method was used, and as we discussed earlier, the 
transmission delay can also be reduced with a deeper NN in the client 
due to potentially smaller intermediate data.

{\bf Cut Index is an important \zztbd{hyper-parameter} for training delay, accuracy and privacy.} {\it Different Cut Indexes bring the following tradeoff: the deeper client NN adds more resource demand at the edge, but reduces the transmission time and enhances attack resilience. On the other hand, a shallower server NN may lead to lower accuracy with SerAVG.}

Furthermore, system architects can combine the approaches mentioned above, e.g., having a moderately deep NN in clients and using the CPA-DC, to find a balance between resource demand and performance.
As in Figure~\ref{accu&resi_MNIST_LENET_aware_03}, with $cut\_index=4$, $DC\_Frequency = 1$ and $loss\_multiplier=0.3$, we still get about $80\%$ attack resilience and more than $90\%$ accuracy.

On the other hand, \zztbd{when} comparing FSL and PSL, we note that FSL has more \zztbd{hyper-}parameters to tune. 
\zztbd{But, i}n all experiments reported in Figure~\ref{multipliers}, when the Cut Index is large, FSL has a better accuracy than PSL. 
So we conclude that \zztbd{a carefully specified way of partitioning the NN can benefit the most when applied in hybrid federated-split learning systems.}


\vspace{-2 mm}
\section{Conclusion}\label{sec:conclusion}
Systems like Federated Learning (FL), Split Learning (SL), and later works aim to fit specific scenarios such as distributed model training and inference.
However, they are not flexible enough to fit some use cases with the recent development in edge and constrained devices. 

In this work, we propose and extensively evaluate Federated Split Learning (FSL), a system for efficient training and inference with high-level privacy for clients' source data.
We present a Client-based Privacy Approach (CPA) for split learning-based systems \zztbd{to provide high attack resilience by adding noise to the intermediate data.}
\zztbd{We also study the training time, accuracy and privacy level of different ways to partitioning a NN in FSL.}
As further works, we aim to research \zztbd{prediction based NN partitioning methods.}

Moreover, comparisons between FSL and existing NN training and inference systems at the edge are carried out, along with explanations of their pros and cons against FSL. 
Among the main analyzed systems, the Parallel Split Learning (PSL)'s principal limitation is a slow and stateful server, and the Federated Reconstruction (FRC) system is inefficient in training time and inference time.

\section*{Acknowledgements}
This work has been supported by National Science Foundation Awards CNS-1908574 and CNS-1908677.

\bibliographystyle{IEEEtran}
\bibliography{IEEEabrv,Tech-Report-FSL}

\vfill



\end{document}